\pdfoutput=1

\documentclass[11pt]{article}

\usepackage[]{acl}

\usepackage{times}
\usepackage{latexsym}

\usepackage[T1]{fontenc}

\usepackage[utf8]{inputenc}

\usepackage{microtype}

\usepackage{inconsolata}

\usepackage{todonotes}
\usepackage{makecell}
\usepackage{amsmath}
\usepackage{bbm}
\usepackage{dsfont}
\usepackage{array}
\usepackage{caption}
\usepackage{subcaption}
\usepackage{enumitem}
\usepackage{amssymb}
\usepackage{pifont}
\newcommand{\cmark}{\ding{51}}%
\newcommand{\xmark}{\ding{55}}%

\usepackage{float}
\usepackage{listings}
\usepackage{setspace}
\usepackage{xcolor}
\usepackage{multirow}
\usepackage{tabularx}
\usepackage{longtable}
\usepackage{soul}

\usepackage{algorithm}
\usepackage{algpseudocode}

\usepackage{array,multirow}

\title{PatentEval: Understanding Errors in Patent Generation}

\author{
  You Zuo\textsuperscript{1,2} \
  Kim Gerdes\textsuperscript{2,3} \
  Éric de la Clergerie\textsuperscript{1} \
  Benoît Sagot\textsuperscript{1} \\
  \textsuperscript{1}Inria, Paris, France  \\
  \textsuperscript{2}Qatent, Paris, France \\
  \textsuperscript{3}LISN, CNRS and University Paris-Saclay, Orsay, France \\
  \texttt{\{you.zuo, eric.de\_la\_clergerie,benoit.sagot\}@inria.fr} \\
  \texttt{gerdes@lisn.fr}
}

\begin{document}
\maketitle
\begin{abstract}

In this work, we introduce a comprehensive error typology specifically designed for evaluating two distinct tasks in machine-generated patent texts: claims-to-abstract generation, and the generation of the next claim given previous ones. We have also developed a benchmark, PatentEval, for systematically assessing language models in this context. Our study includes a comparative analysis, annotated by humans, of various models. These range from those specifically adapted during training for tasks within the patent domain to the latest general-purpose large language models (LLMs). Furthermore, we explored and evaluated some metrics to approximate human judgments in patent text evaluation, analyzing the extent to which these metrics align with expert assessments. These approaches provide valuable insights into the capabilities and limitations of current language models in the specialized field of patent text generation.
\end{abstract}

\section{Introduction}
A patent is a legal instrument that grants inventors or entities exclusive rights over their invention for a designated period. This exclusivity is said to stimulate innovation by safeguarding the intellectual property of the inventors. 
Patent drafting refers to the process of writing a detailed description of an invention in a legal document that meets the requirements of patent law. It is a complex and time-consuming task that requires a thorough understanding of the invention and the relevant patent laws, and the cost of obtaining a patent can be significant \cite{costpatent}, with the drafting process being the biggest part of the cost. 

With the progression of deep learning technologies, a multitude of complex challenges in the patent domain have been ameliorated. Computational techniques have notably enhanced patent prior art searches \cite{risch2020patentmatch, buckley2021patentexplorer, VOWINCKEL2023102192} and facilitated efficient patent classification \cite{lee2019patentbert, huang2019hierarchical}.

However, the generation of patent texts and the assessment of the quality of text produced by neural models remain underexplored areas of research. This can be attributed to the exigent demands for precision and accuracy within the legal domain, and it is clear that the evaluation of machine-generated patent texts necessitates an extensive domain-specific acumen, attainable only by experts within the field. This additional layer of complexity poses significant hurdles for researchers from allied disciplines, seeking to venture into and assess their contributions to this niche, unlike in more generic text applications such as machine translation or generic dialog systems. Nonetheless, recent strides made by large language models, such as OpenAI’s GPT-3.5, GPT-4 \cite{openai2023gpt4}, and other open-source variants like llama2 \cite{touvron2023llama} and Falcon \cite{refinedweb}, have demonstrated promising capabilities in generating high-caliber legal texts \cite{choi2023chatgpt}. These developments signal a promising horizon for enhanced performance and nuanced evaluation in the domain of patent text generation.

In our study, we focus on evaluating and understanding the quality of patent text generation by various language models. We have developed \textbf{PatentEval}, a benchmark annotated by human experts, tailored for assessing language models of different sizes and capacities. This includes pairwise comparisons and detailed analysis of error types in each output. Our goal is to narrow the gap between human-written and machine-generated patents, offering a clearer view of the potential uses of large language models in this field. 

\section{Preliminaries and Background}
A patent is a structured document that typically includes several sections, such as a title, abstract, background, brief summary of the invention, detailed description, one or more claims, drawings, and classification information, among others. In this study, our primary focus is on the generation of patent abstracts and claims.

The abstract of a patent is a concise summary that offers a straightforward overview of the invention's main features. It is typically used for informational and search purposes, helping individuals quickly understand the essence of the patented technology without delving into the detailed description found in the patent specification.

Patent claims stand as the cornerstone of a patent document. Claims meticulously define the specific features and associated rights of an invention. Written in a unique combination of legal jargon and patent-specific language, these claims serve to concisely and unambiguously detail the novel elements of an invention. This could relate to its construction, composition, or operational methodology. These claims set the boundaries for what others can or cannot do without permission from the patent holder. In this sense, writing the claims is a strategic choice of the patent council that depends on outside, e.g. economic factors, and is not as a whole automatable. Nonetheless, a system that proposes subsequent claims can be of great use to the council to ensure the quality and completeness of the claim set.

\begin{table}
\small
\centering
\begin{tabular}{>{\raggedright\arraybackslash}p{0.52\linewidth} >{\raggedright\arraybackslash}p{0.40\linewidth}}
\hline
\textbf{Independent claim} & \textbf{Dependent claim}\\
\hline
1. A lighted pencil, comprising:\newline a pencil shaft; and\newline a light attached to the pencil shaft. & 
2. The lighted pencil of claim 1, wherein the light is removably attached to the pencil shaft.\\
\hline
\end{tabular}
\caption{Examples\protect\footnotemark of independent/dependent claim. }
\label{indep_dep}
\end{table}
\footnotetext{Examples taken from: \url{https://www.wipo.int/edocs/mdocs/aspac/en/wipo_ip_mnl_3_18/wipo_ip_mnl_3_18_p_5.pdf}}

In addition, claim dependency is another variable to look at (table~\ref{indep_dep} shows examples of both types): 

\begin{itemize}[noitemsep,topsep=0pt]
    \item \textbf{Independent Claims:} These claims encapsulate the invention's core features without referencing other claims. They represent the invention's essence autonomously. According to the United States Patent and Trademark Office (USPTO) drafting regulations, a patent can have multiple independent claims embodied in the invention. 
    \item \textbf{Dependent Claims:} These claims reference and build upon one or more prior claims, either independent or dependent. The dependent claims include everything recited in their independent claims. They augment the independent claims by introducing extra details, variants, or features, resulting in a more circumscribed protection ambit as they adopt the restrictions of the claims they refer to.
\end{itemize}

\section{Related Work}
Recent advancements in natural language generation (NLG) have seen significant progress across various domains, yet generating and evaluating patent texts remains a challenge due to the intricacies of legal knowledge.

Initial strides in this field were made by the PatentTransformer project \cite{lee2020patent}, which explored adapting the GPT-2 model \cite{radford2019language} to generate patent claims, aiming to assist patent writers with an "augmented inventing" tool. A subsequent version of PatentTransformer \cite{lee2020patenttransformer} expanded this capability to generate different patent sections from given parts (e.g., converting an abstract into a title or claim).

To evaluate the effectiveness of these generated claims, \cite{lee2019measuring} fine-tuned a Bert model \cite{devlin2018bert} for binary classification, assessing the relevance of consecutive claim segments. Building on this, \cite{lee2020measuring} developed a two-Transformer model framework for quality control in patent text generation, proposing an "auto-complete" feature to facilitate idea exploration from existing patents. Additionally, their study \cite{lee2020prior} investigated the origins of generated content by applying prior-art search techniques to the training data, laying groundwork for future assessments of text novelty in patents.

The IBM research team introduced the Patent Generative Transformer (PGT) \cite{christofidellis2022pgt}, enhancing the GPT-2 model for multifaceted tasks in the patent domain, such as part generation, text infilling, and coherence checking. To assess PGT, they employed methods like semantic similarity comparison, expert evaluations,\footnote{Their evaluation, however, was limited to 44 patents in the chemistry domain and remains unpublished.} and analysis of the model's zero-shot performance on novel generation tasks. More recently, \cite{lee2023evaluating} experimented with various sizes of PatentGPT-J, rooted in GPT-J \cite{gpt-j}, and introduced a novel metric that gauges the efficiency of language models in generating patent claims by quantifying the reduction in keystrokes for autocomplete functions.

However, the variation in evaluation methodologies and datasets across studies, including those focusing on patent summarization or claim generation, makes it difficult to compare results consistently.

Several datasets and benchmarks have been developed for patent-related tasks. The Big Patent dataset \cite{sharma2019bigpatent} focuses on patent summarization and includes about 1.3 million U.S. patent documents sourced from the Google Patent Public Datasets via BigQuery. Its text coherence and abstractiveness were evaluated using n-gram occurrence rates and entity distribution metrics. The Harvard USPTO Dataset (HUPD) \cite{suzgun2022hupd} is another crucial resource, comprising English-language utility patent applications filed with the USPTO between 2004 and 2014. It features benchmarks for binary patent decision classification, multi-class IPC/CPC classification, masked language modeling, and abstractive summarization, with task-specific metrics like ROUGE for summarization. 

Furthermore, \cite{casola2022summarization} emphasized the challenges in ensuring factual consistency in patent texts, suggesting alternative evaluation methods like QAGS \cite{wang2020asking} and FactCC \cite{kryscinski2019evaluating}, and FactGraph \cite{ribeiro2022factgraph}, aligning also with our research focus.

Recent advancements in large language models (LLMs), exemplified by OpenAI’s GPT-3.5, GPT-4 \cite{openai2023gpt4}, and other open-source alternatives such as Llama2 \cite{touvron2023llama} and Falcon \cite{refinedweb}, have showcased their capabilities for zero-shot learning and their adeptness at handling a wide array of tasks when provided with straightforward instructions.

In light of these developments, our study aims to comparatively assess the performance of these models, with a particular emphasis on shared tasks like claims generation and abstract generation. We benchmark these models against their contemporary generative counterparts using a dedicated evaluation dataset. Moreover, we delve into a nuanced analysis of their outputs by investigating the characteristics and distribution of the errors they produce.

\section{Tasks and Criteria}

\subsection{Tasks}
To evaluate the capabilities of different models more comprehensively, we selected two tasks targeting distinct generation content formats: one for abstracts and another for claims. We intentionally chose one generation task as a summarization task given input (claims2abstract), while the other lacks a standardized answer (next claim generation). This design ensures a more nuanced assessment of the models' capabilities.

\noindent
\textbf{Claims2Abstract} \quad
In patents, a claim is a legally binding description defining the patent's protection written in a formal legal style. A patent often has multiple claims specifying its scope. Conversely, a patent's abstract offers a brief summary of its technical details and implications. Hence, the claims-to-abstract task is highly relevant in the context of patent summarization, where the objective is to transform the juridical language of claims into more generic and concise abstracts.

In the Claims2Abstract task, the input consists of the full set of claims.
The objective is to generate an abstract that encapsulates the patent's main elements.


\noindent 
\textbf{Next Claim Generation} \quad
Previous studies \cite{lee2020patent, lee2020patenttransformer, lee2023evaluating} have approached claims generation as a means of "augmented inventing," aiming to develop a tool that assists human patent practitioners by providing autocomplete suggestions during the drafting process. These works aimed to let large language models generate claims from scratch or with minimal input, with evaluation metrics concentrated on word-level or span-level aspects of the claims. Instead of incremental generation or evaluation, we focus on producing the entire subsequent claim in one go. This approach not only tests the models' capabilities in a more holistic manner but also aligns more closely with the practical needs of patent drafting, where each claim needs to be fully formulated and coherent in itself.


In the generation phase of our models, we varied the input by providing either the first claim alone (claim 1), the first and second claims together (claims 1-2), or the first three claims (claims 1-3). The objective for the model in each scenario is to generate the next sequential claim.

One of the key evaluation criterion is the model’s ability to produce a subsequent claim that not only follows logically but also matches the dependency type (independent or dependent) of the corresponding original claim in the patent. This means if the original subsequent claim in the patent is an independent claim, the model-generated claim should also be independent, and similarly for a dependent claim. This approach ensures that the generated claim maintains the same structural and legal relationship as the original set of claims, and easier for us to do the evaluation and comparison with human-drafted patents.

\subsection{Typology of Errors}
While prior studies have delved into specific aspects of generated content, such as relevance among spans of claims \cite{lee2019measuring} or the semantic similarity between generated and actual components \cite{christofidellis2022pgt}, the process of patent drafting encompasses a broader set of criteria. Beyond ensuring syntactic accuracy and semantic relevance, the content must be patentable under prevailing regulations, avoid the use of prohibited terms specific to patent language, and the patent application should articulate the invention with both clarity and comprehensiveness.

To better understand and categorize these multifaceted errors, we established a typology based on the issues observed in outputs from various models. Our error types also refer to the guidelines from the second edition of the WIPO Patent Drafting Manual \cite{WIPOPatentDrafting2022}.

As we use USPTO data, we have based this work primarily on the USPTO's patent drafting standards as well. More detailed explanations of each error type with examples are demonstrated in appendix \ref{error_example}.

\subsubsection{Abstract Generation}
A good abstract gives a quick overview of the invention’s key technical points. It’s often the first thing seen on a patent’s first page and is used in search databases, guiding automated search tools with its keywords. Therefore, the abstract should be short but also accurate, offering a clear snapshot of the invention’s details.
We thus summarize the following dimensions for errors in abstract drafting:

\noindent
\textbf{Grammatical Errors:} Occurrences of incorrect grammar, punctuation, or sentence structure, including hallucinated repetitive sequences produced by language models.

\noindent
\textbf{Irrelevant Content:} Introducing content that deviates or digresses from the primary subject matter of the patent claims.

\noindent
\textbf{Incomplete Coverage:} \citep[p.~106]{WIPOPatentDrafting2022} Occurrences where the abstract omits essential components or concepts, failing to encapsulate all key points from the patent claims, especially the main (first independent) claim.

\noindent
\textbf{Overly Wordy or Lengthy:} \citep[p.~107]{WIPOPatentDrafting2022} Abstracts falling into this error type are not succinct, containing unnecessary details. Jurisdictions often impose word limits on abstracts — for example, in many English-speaking countries, abstracts are typically restricted to 150 words.

\noindent
\textbf{Contradictory Information:} Instances when the abstract introduces factual details that contradict the content found in the original claims.

\noindent
\textbf{Unclarity:} The abstract contains vague or ambiguous descriptions, making it difficult to grasp the intended message or details.

\noindent
\textbf{Ineffective Summarization:\footnote{While instances exist where the first claim of a patent is minimally modified and used as the abstract, such practice is not advisable. We include this as an error for the reason that we want to have models that can provide a clear and comprehensive summary of the invention as a whole, rather than merely replicating the language of the claims.}} Relates to abstracts that inadequately summarize the invention, often replicating one or more of the claims verbatim instead of providing a concise and comprehensive overview of the patent.

\subsubsection{Claim Generation}
Patent claims are structured sentences that distinctly describe the invention seeking protection. To be patentable, these claims must show novelty, be non-obvious compared to existing "prior art," and have practical application. Due to the extensive time and resources required for prior art searches and verification, our study focuses 
on the inherent structure that can be evaluated without access to exterior databases (patents, scientific articles, and so on).
We leave the evaluation of novelty and non-obviousness of generated claims for future work.

Given the complexity and stringent drafting rules of claims compared to other patent sections, we have developed a detailed typology of error types: 

\noindent
\textbf{Grammatical Errors:}
\begin{itemize}[noitemsep,topsep=0pt]
\renewcommand\labelitemi{--}
    \item \textbf{Grammatical Inaccuracy:} Misuse of grammar and hallucinated repetitive sequences produced by language models.
    \item \textbf{Punctuation Discrepancy:} \citep[p.~41]{WIPOPatentDrafting2022} Incorrect or inconsistent use of punctuation marks, deviating from standard patent drafting conventions.
\end{itemize}

\noindent
\textbf{Formatting Errors:}
\begin{itemize}[noitemsep,topsep=0pt]
\renewcommand\labelitemi{--}
    \item \textbf{Claim Numbering Error:} Incorrect or inconsistent numbering of claims.
    \item \textbf{Preamble\footnote{The preamble of a patent claim provides an introductory description of the invention, setting the context or intended use. For example, in a claim for a new type of smartphone, the preamble might state: "A communication device designed for handheld use, ..." to establish the device's general category and purpose.} Inconsistency Error:} \citep[p.~39]{WIPOPatentDrafting2022} Inaccurate reflection of subject matter in the preamble, disrupting the conceptual flow between independent and dependent claims.
    \item \textbf{Transitional Phrase\footnote{Transitional phrases in patent claims, such as "comprising," "consisting of," and "consisting essentially of," define the scope of the invention. For example, a claim stating "A device comprising A, B, and C" allows for additional elements beyond A, B, and C, whereas "A device consisting of A, B, and C" restricts the invention to only those three components.} Error:} \citep[p.~40, 43-44]{WIPOPatentDrafting2022} Improper use of transitional phrases, impacting the scope of the claim.
    \item \textbf{Claim Body Disconnection:} \citep[p.~41]{WIPOPatentDrafting2022} Presence of fewer than two elements or a lack of a coherent, logical connection between listed elements in the claim body.
\end{itemize}

\noindent
\textbf{Dependency Errors:}
\begin{itemize}[noitemsep,topsep=0pt]
\renewcommand\labelitemi{--}
    \item \textbf{Non-compliant Dependency with instruction:} Dependency of the claims not matching the required dependency as instructed.
    \item \textbf{Dependency Clarity Error:} \citep[p.~50-60]{WIPOPatentDrafting2022} Utilization of unclear multiple dependencies or an incorrect singular dependency.
    \item \textbf{Broad Scope Dependent Claims:} \citep[p.~52]{WIPOPatentDrafting2022} Dependent claims that insufficiently narrow the scope of the independent claim they depend on.
    \item \textbf{Insufficient Differentiation of Independent Claims: } \citep[p.~48-50, 85]{WIPOPatentDrafting2022} Independent claims that cover the same or similar scope as previous claims.
\end{itemize}

\noindent
\textbf{Clarity Errors:}
\begin{itemize}[noitemsep,topsep=0pt]
\renewcommand\labelitemi{--}
    \item \textbf{Vagueness:} \citep[p.~24, 80-82]{WIPOPatentDrafting2022} Usage of ambiguous, vague, or relative terms or expressions that render the claim's scope indefinite.
     \item \textbf{Antecedent Reference Errors:} \citep[p.~42]{WIPOPatentDrafting2022} Failure to provide a clear antecedent basis for each term.
    \item \textbf{Terminological Inconsistency:} \citep[p.~43, 79]{WIPOPatentDrafting2022} Use of multiple terms or different reference numerals for the same element. 
    \item \textbf{Wishful Claiming:} \citep[p.~68-69]{WIPOPatentDrafting2022} Claims that express objectives without concrete methods, leading to speculative or abstract language.
\end{itemize}

\noindent
\textbf{Brevity Errors:}
\begin{itemize}[noitemsep,topsep=0pt]
\renewcommand\labelitemi{--}
    \item \textbf{Verbose Redundancy:} Excessive wordiness without adding substantive content.
    \item \textbf{Sub-Optimal Claim Structure:} \citep[p.~47, 49]{WIPOPatentDrafting2022} Claims with complex language that could be more clearly expressed as multiple, simpler claims.
\end{itemize}

\noindent
\textbf{Content Relevance Errors:}
\begin{itemize}[noitemsep,topsep=0pt]
\renewcommand\labelitemi{--}
    \item \textbf{Irrelevant Matter Introduction:} Introduction of matter unrelated to the disclosed embodiments, potentially broadening the claim beyond the invention's scope.
\end{itemize}

\noindent 
\textbf{Effectiveness Error:}
\begin{itemize}[noitemsep,topsep=0pt]
\renewcommand\labelitemi{--}
    \item \textbf{Contradictory Claims:} Claims that conflict with previous claims or do not follow a logical flow themselves.
    \item \textbf{Non-Distinctive Claim Repetition:} Claims that lack effectiveness, primarily repeating content from earlier claims without adding new scope or detail.
\end{itemize}

\section{Dataset Creation}

\subsection{Data Selection}
Our experimental dataset used for constructing input during inference originates from the Harvard USPTO Dataset (HUPD) \cite{suzgun2022hupd}. This comprehensive corpus encompasses English-language utility patent applications submitted to the USPTO spanning January 2004 through December 2018.

Given that HUPD comprises both granted and rejected patent applications, we selectively included only those patents that had been granted, ensuring the inclusion of high-quality patent text. Additionally, we eliminated entries bearing "(canceled)" claims, as they tend to be non-informative for patent drafting and could introduce undesirable noise into the dataset. Subsequently, we formed evaluation datasets by randomly sampling 400 granted patents – equating to 50 from each of the eight primary IPC \footnote{\url{https://www.wipo.int/classifications/ipc/en/}} sections – from the years 2017 and 2018, ensuring a balanced representation across all patent domains. We use claims from these patents for constructing inputs of models for both tasks of Claims2Abstract and next-claim-generation.

\subsection{Models under Evaluation}
\begin{table*}[ht]
\small
\centering
\begin{tabular}{lrccc}
\hline 
\textbf{Models} &\textbf{Size} & \textbf{Context length} & \textbf{Claims2Abstract} & \textbf{next-claim} \\ 
\hline
PatentTransformer & 1.5B & 1024 & \cmark & \cmark \\ 
\hline
PGT & 1.5B & 1024 & \cmark & \xmark \\ 
\hline 
HUPD T5-Small & 60M & 512 &  \cmark & \xmark \\ 
\hline
PatentGPT-J  & 1.6B & 2048 & \xmark & \cmark\\
\hline
Falcon  & 7B, 40B & 2048 & \cmark & \cmark \\ 
\hline
Llama 2 & 7B, 13B, 70B & 4096 & \cmark & \cmark\\ 
\hline
gpt-3.5-turbo-0613  &  & 4097 & \cmark & \cmark \\ 
\hline
\end{tabular}
\caption{\label{info_model} Overview of language models selected for patent generation evaluation. }
\end{table*}

Table~\ref{info_model} compares basic information and the tasks supported among the selected models for our evaluation. In order to provide a comprehensive assessment, our selection encompasses both specialized models \cite{lee2020patenttransformer, christofidellis2022pgt, suzgun2022hupd, lee2023evaluating} designed explicitly for patent-related tasks and the latest Large Language Models (LLMs), such as Llama 2 \cite{touvron2023llama} and Falcon \cite{refinedweb} of various sizes. The majority of these models are built upon decoder-only architectures, whereas the model introduced in \cite{suzgun2022hupd} adopts an encoder-decoder architecture based on T5 \cite{raffel2020exploring} for patent summarization tasks.

In addition to the open-source models tailored specifically for patents, we incorporate OpenAI's latest fixed version GPT-3.5, \texttt{GPT-3.5-turbo-0613}, into our evaluation. To minimize randomness, we set the temperature parameter to 0, while keeping default values for other hyperparameters. 

During inference, each relevant model produces a single output for each of the 400 chosen patents across the two tasks. Detailed model inferences can be found in Appendix \ref{sec:models}.

\subsection{Annotation Data}

Our human evaluation concentrated on domains where we have the most expertise, analyzing 50 patents each from domain A (human necessities) and domain G (physics)\footnote{Model outputs for the other six domains are reserved for future studies and can be accessed at \url{https://github.com/ZoeYou/PatentEval}.}. In our next-claim-generation task, we particularly examined whether the models could accurately generate claims according to the required dependency criteria. Due to a limited number of independent subsequent claims in these domains, we also included additional examples from other domains, adding eight more instances to our analysis.

Some of the primary objectives of our research include 1) assessing the capabilities of various models in generating patent texts, and 2) exploring whether human evaluators have a preference for contents generated by humans or machines.

To achieve these objectives, our annotation process involved comparative evaluations. Annotators were presented with two types of paired outputs for each input claim or set of claims:
\begin{enumerate}[noitemsep,topsep=0pt]
    \item A comparison between two different models $(\text{model}_{11}, \text{model}_{12})$;
    \item A juxtaposition of a model's output against the original abstract or subsequent claim $(\text{model}_{21}, \text{original abstract/next claim})$.
\end{enumerate}
Models for each pair were randomly selected to ensure variety and prevent bias, with $\text{model}_{11}$ and $\text{model}_{12}$ always being different, and $\text{model}_{21}$ chosen independently from them.

\section{Results and Analysis}

The annotation process involved two primary annotators: a seasoned patent lawyer with over 15 years of experience in relevant domains, and a PhD student. In cases of disagreement, a third expert was consulted to reach consensus. Our analysis is structured around addressing three key questions:

\begin{figure*}[t!]
\hspace*{-40mm} 
    \centering
    \begin{subfigure}[t]{0.49\textwidth}
        \centering
        \includegraphics[height=2.4in]{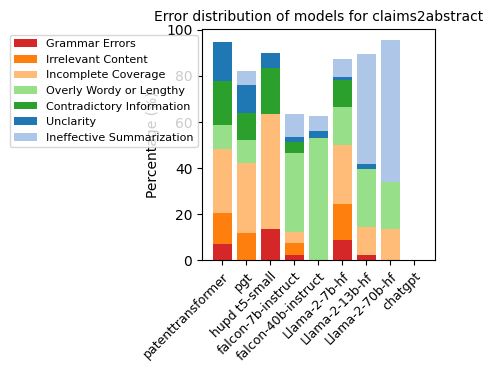}
    \end{subfigure}%
    ~ 
    \begin{subfigure}[t]{0.48\textwidth}
        \centering
        \includegraphics[height=2.5in]{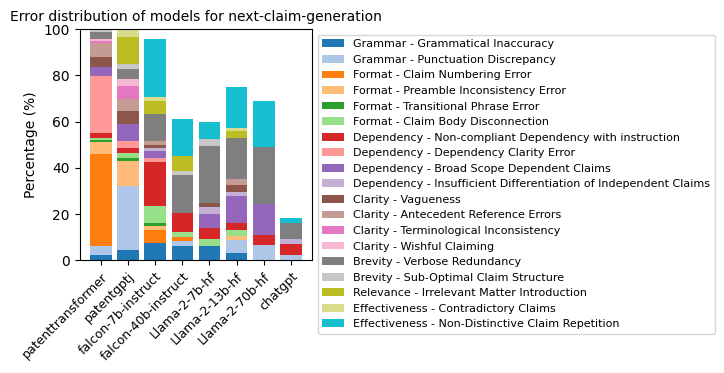}
    \end{subfigure}
    \caption{Error distribution of different models evaluated on two tasks.}
    \label{error_dist}
\end{figure*}

\subsection*{Q1: What are the error distributions of each model?}
Figure~\ref{error_dist} shows the error distribution in 416 pairs for both task outputs of different models. The distance of the bars to the top represents the proportion of error-free outputs. Below that, the segmented bars illustrate the percentages of specific error types occurring. Notably, ChatGPT excelled in minimizing errors in both quantity and diversity.

In the Claims2Abstract task, ChatGPT demonstrates exemplary performance, flawlessly executing the task without any errors. Among models within the large language model (LLM) family, a discernible trend emerges, wherein larger models tend to exhibit a reduction in the variety of errors. For instance, both Llama2-7b and Falcon-7b exhibit basic errors, including grammatical mistakes, hallucinations, and other types of errors. In contrast, Falcon-40b and Llama2-70 are more prone to errors related to the coverage of the invention's scope. Falcon-40b frequently generates repetitive and overly verbose abstracts, while Llama2-70 struggles to fully capture the scope of the invention.

The HUPD T5-small model, constrained by its limited encoder and decoder context lengths, often overlooks essential components of claims and incorporates non-factual information. However, its fine-tuning on actual claims-to-abstract data ensures that the abstracts it generates maintain close relevance to the given claims. In contrast, other models exhibit a variety of errors to differing extents.

In the next-claim-generation task, improper punctuation usage is common across models, often influenced by the non-standard punctuation usage in the input claims. Unlike ChatGPT and Llama2-70b, which manage to avoid making grammatical errors, the majority of models grapple with grammatical inaccuracy in their drafting. PatentTransformer, PatentGPT-J, and Falcon-7b, in particular, struggle with aligning with the claims formatting, where PatentTransformer and PatentGPT-J also have trouble maintaining consistent antecedent referencing and avoiding preamble inconsistency, issues that frequently co-occur.

Large language models (LLMs) often demonstrated a tendency to rephrase or repeat previous claims without enhancing specificity in the generated dependent claims, thereby disrupting the logical flow of the claims' scope. This issue was particularly pronounced among Llama-2 models. Additionally, Falcon models frequently failed to adhere to the specified dependencies outlined in instructions, resulting in compromised coherence in the generated claims. Moreover, PatentTransformer exhibited a propensity to misnumber claims or reference non-existent prior claims, further undermining the integrity of claim sequences.

Excluding ChatGPT, Llama2-7b, Llama2-70b, and PatentTransformer, models at times generated content that was either irrelevant or factually incorrect, highlighting the significant challenge of achieving accuracy in patent claim generation.

\begin{figure}[t!]
\hspace*{-35mm}
    \centering
    \begin{subfigure}[t]{0.5\textwidth}
        \centering
        \includegraphics[height=2.1in]{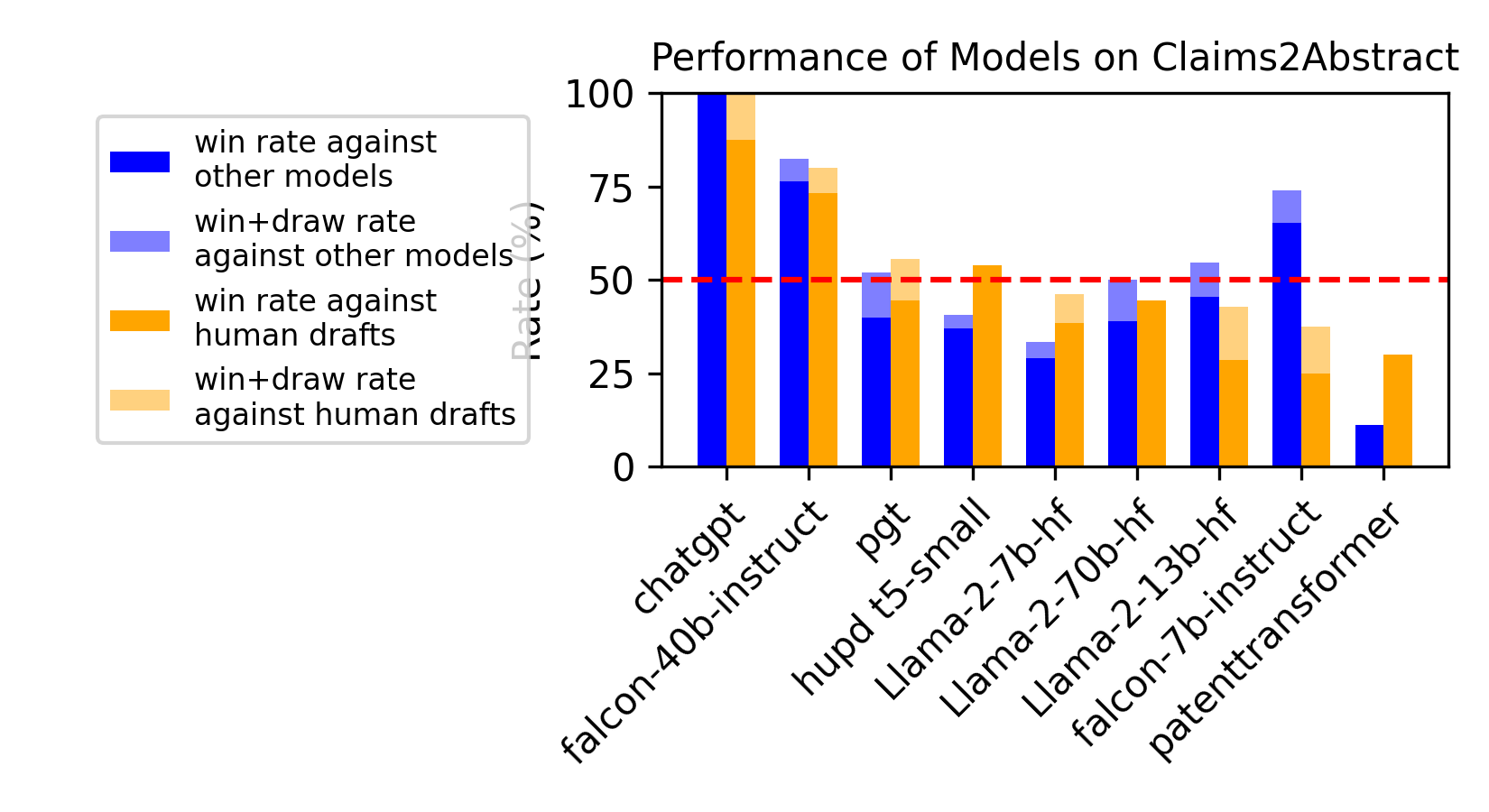}
    \end{subfigure}%
    \\
    \begin{subfigure}[t]{0.5\textwidth}
\hspace*{-20mm}
        \centering
        \includegraphics[height=2.1in]{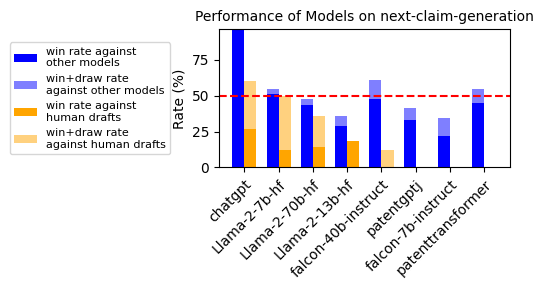}
    \end{subfigure}
    \caption{Combined win and draw rates of different models evaluated on two tasks (compared to original abstract/claim or other models in the sampled dataset).}
    \label{winning_rate}
\end{figure}

\subsection*{Q2: Do machines perform better in patent drafting?}
We conducted a statistical analysis of pairwise annotations, calculating the win and draw rates of each model against others in the sampled datasets and against human-drafted outputs. The results are depicted in Figure~\ref{winning_rate}. In the bar plots, the deeper color in each bar represents the win rate, while the draw rate starts from the win rate value, resulting in the top of the bar representing the sum of win and draw rates.

In the claims2abstract task, large language models (LLMs) such as ChatGPT, Falcon-40b, and fine-tuned models like pgt and HUPD T5-small produced abstracts preferred over or performed equally well as human-written ones, with a winning rate exceeding 50\%. Similarly, in the task of next-claim-generation, ChatGPT's outputs were favored over or performed comparably to human-generated claims more than 50\% of the time, while Llama2-7b matched human preference rates at 50\%.

These findings suggest that, despite their imperfections, these large language models (LLMs) possess significant potential and can offer valuable assistance in patent drafting tasks.

\subsection*{Q3: When do human drafters perform worse than certain models?}

\begin{figure}[t!]
    \centering
    \begin{subfigure}[t]{0.5\textwidth}
        \centering
        \includegraphics[height=1.7in]{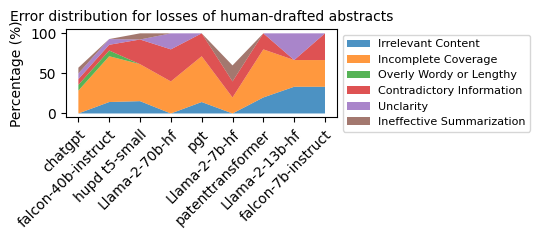}
    \end{subfigure}%
    \\
    \begin{subfigure}[t]{0.5\textwidth}
        \centering
        \includegraphics[height=1.7in]{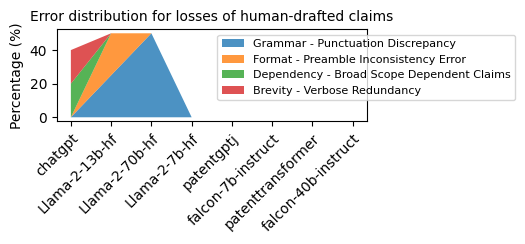}
    \end{subfigure}
    \caption{Error distribution for losses of human-drafted abstracts/claims. (from left to right lies the models in the order that human drafts have a lower loss rate.)}
    \label{fig:hints}
\end{figure}

To gain deeper insight into why certain models outperform human-drafted abstracts or claims, we visualize the types of errors present in human-drafted contents when they perform worse than machine-generated ones. As depicted in Figure~\ref{fig:hints}, human-drafted abstracts consistently fall into the category of "incomplete coverage" when they perform worse than machine-generated abstracts. This phenomenon can be attributed to strategic choices made by patent drafters. It's important to note that an abstract is not part of the claims or specification as filed; instead, it provides information for search purposes later, often by adversaries. Consequently, patent drafters may intentionally bias their abstracts to render them more or less detectable. By including keywords present in other parts of the claim tree besides the first independent claim, one increases the discoverability of the patent application. Conversely, by merely copying claim 1 without restituting other keywords, one decreases detectability. This bias may explain why claims2abstract models indicate that humans perform poorly on the surface.

As for other main errors made by humans, such as including irrelevant content and contradictions, this could be attributed to the timing of abstract drafting. Abstracts are typically drafted when a patent application is filed. However, in our case, we selected granted patents, which represent the final examined and amended version of the patent application. During the amendment process, while claims or other parts of the patent application may be modified, the abstract remains as originally filed unless formally amended by the examiner with the applicant's approval.

\section{Evaluation of Metrics}
\begin{table}[ht!]
\small
\centering
\begin{tabular}{p{1cm}|lc}
\hline 
\textbf{Task} & \textbf{Metric} & \textbf{Kendall's Tau}\\ 
\hline \hline
\multirow{6}{*}{abstract}   & SemSim (without fine-tuning) & .2562 \\
                                & SemSim (fine-tuned on IPC) & .2662 \\
                                 \cline{2-3}
                                 & \textbf{Terms Coverage} & \textbf{.2865} \\ 
                                 & N-grams Coverage & .1767  \\ 
                                 \cline{2-3}
                                 & FactGraph & .0653\\ 
                                 & QAFactEval & .2507\\
\hline \hline
\multirow{6}{*}{claim} & \textbf{Rule-based checker} & \textbf{.4120} \\ \cline{2-3}
                            & SemSim (without fine-tuning) & .1278 \\ 
                            & SemSim (without fine-tuning)* & .2848\\  
                            & SemSim (fine-tuned on IPC) & .0249\\
                            & SemSim (fine-tuned on IPC)* & .2568\\ \cline{2-3}
                            & EntityGrid  & .0309\\
\hline
\end{tabular}
\caption{\label{kendall's tau}Kendall's tau correlation of evaluation metrics with manual annotation. (The * in the next claim generation task indicates that the metric score is weighted by the rule-based checker score.)}
\end{table}

In this section, we evaluate various metrics to assess patent generation, focusing on their alignment with human judgments. Detailed information on these metrics is available in Appendix~\ref{Metrics}. This assessment aids in understanding the extent to which automated metrics can accurately mirror human evaluations in the context of patent text generation to facilitate the future evaluation of patent generation. 

Using all our human annotations from pairwise comparisons across two tasks, we assess the metrics' performance. We apply these metrics to the same sets of input and output that human annotators reviewed in pairwise comparisons. Outputs are scored by each metric, with the higher-scoring output ranked as "1" (preferred) and the lower-scoring as "2". In cases where a metric assigns identical scores to both outputs, the outputs are ranked equally as (1, 1). Ultimately, we compile a list of pairs ranked in this manner, facilitating comparison with the lists generated by human annotators.


Our table~\ref{kendall's tau} below highlights the correlation of each metric with manual annotation for our main tasks, indicating their effectiveness in mirroring human judgment.

In the task of abstract generation, we observed that the coverage of technical terms from input claims exhibited a reasonable correlation with human evaluations, achieving a score of 0.2865. This was closely followed by a semantic similarity metric (SemSim), with a score of 0.2662, utilizing a BERT-for-patent \cite{SrebrovicYonamine2020} model fine-tuned on the International Patent Classification (IPC) task. We also explored two other existing metrics designed for generic text summarization factuality evaluation: FactGraph \cite{ribeiro2022factgraph}, which integrates information extracted from both AMR graphs and text, and QAFactEval \cite{fabbri-etal-2022-qafacteval}, a QA-based model. 

However, FactGraph's performance was subpar, likely due to its inability to effectively extract AMR information from patent texts, given their complex structure and intricate relationships among entities. Surprisingly, QAFactEval, demonstrated a high correlation with our human judgments. This underscores the effectiveness of the strategy employed by QA models, wherein a question is posed based on selected information and answered using input text.

In the realm of claim generation, we introduced a heuristic method specifically designed to identify rule-based errors, effectively converting instances of errors into a normalized score. Notably, this metric demonstrated a superior correlation with human judgments compared to other methods tested. This outcome underscores the potential benefit of incorporating additional patent drafting rules into the metric, thereby refining and enhancing its accuracy further. To leverage the insights gained from the rule-based checker, we utilized its scores to weight (multiply) the scores assigned by other metrics. The rationale behind this approach is intuitive: if a drafted claim contains numerous basic errors identifiable by simple rules, it is likely to deviate significantly in quality. The results presented in the table substantiate this hypothesis.

Furthermore, we observed that the fine-tuned model on the International Patent Classification (IPC) performed worse than the original model for claim generation. This observation may suggest that a good next claim does not need to strictly adhere to the same IPC category as its previous claims and should explore broader scope. Further analysis is warranted to explore the implications of this finding.





\section{Conclusions}

This study marks a pivotal advancement in generating and evaluating patent texts, especially abstracts and claims, created by diverse language models. We aimed to explore the potential of LLMs in patent drafting. Our investigation reveals the strengths of certain LLMs in generating quality patent texts and also identifies common errors and their frequencies. These insights lay the groundwork for future progress in this area, informing both the enhancement of existing models and the incorporation of AI into patent drafting practices.


\newpage
\section{Ethical Considerations and Limitations}


This study is confined to the claims-to-abstract as well as the next-claim-generation tasks for patent generation. We can easily extract claim-abstract pairs from the dataset, as these components are independently submitted by applicants and subsequently published by patent offices. 
Claims have to be numbered and are thus easily extractable. However, the main body of the patent application, known as the "description", poses more significant challenges due to its length, often extending to dozens of pages, and its mostly unstructured nature.

Given the current capabilities of Large Language Models, an effective strategy might involve segmenting the patent description into smaller, more manageable sections. These sections, which could include areas like "Background Art", 'Problem Statement', or 'Definitions of Technical Terms', could then be generated using specific models designed for their particular characteristics. Existing work is currently underway to construct expansive datasets of patent text with this level of granular division \cite{liu2023annotating}, but the efficacy of this approach is yet to be definitively proven. It remains an open question whether the performance observed in more standardized sections will carry over to these less regimented areas.

It is important to acknowledge that the scope of this study is restricted to English language patent applications within the USPTO database. When considering other prominent patent languages, all except Chinese offer significantly smaller corpora. This size discrepancy raises uncertainty around the potential to replicate our findings in these languages, given that Language Models tend to demonstrate reduced effectiveness when applied to languages other than English.

Another potential limitation of our study is the inherent bias in comparing different models, particularly since the most recent Large Language Models (LLMs) might have already been exposed to extensive text data, including patents, during their pre-training phase. There's a substantial likelihood that these models have been trained on USPTO patent documents available in open-source datasets. This overlap could inadvertently skew the performance of these models, as they might not be generating content based on learned patterns but rather recalling previously seen data. To address this, future research could implement methods like those proposed by \cite{shi2023detecting} to identify and mitigate potential data pollution. This would involve a thorough examination of the training datasets of these models to ensure the novelty and authenticity of their content generation capabilities, especially in specialized domains such as patent generation.

\section*{Acknowledgements}
We are grateful to all anonymous reviewers for
their valuable comments that have helped to improve this paper. We are also thankful to François Veltz and Lufei Liu for the discussion and valuable feedback on our annotation guidelines. This work was partly funded by the last author’s chair in the PRAIRIE institute funded by the French national agency ANR as part of the “Investissements d’avenir” programme under the reference ANR-19-P3IA-0001.

\bibliography{anthology,custom}

\appendix
\section{Model Details}
\label{sec:models}
This section outlines the specifics of the models used in the PatentEval tasks. Our selection includes transformer architecture-based language models fine-tuned for patent-related tasks, as well as several high-capacity Large Language Models (LLMs).

\begin{itemize}    
    \item \textbf{PatentTransformer} \cite{lee2020patenttransformer}: This model, akin to GPT-2 in architecture, was trained from scratch with approximately 390 million patents from Google Patents Public Datasets on BigQuery (1976-2016).\footnote{\url{https://console.cloud.google.com/bigquery?p=patentspublic-data}} The model was trained on patent text-to-text generation flow (from a few words to a title, the title to an abstract, the abstract to an independent claim, and the independent claim to multiple dependent claims. The text flow can go backward as the relations are trained bidirectionally in their training process.) We used the M2 checkpoint from their GitHub \footnote{\url{https://github.com/jiehsheng/PatentTransformer/blob/master/v2/PatentTransformer_v2.ipynb}} without altering other hyperparameters.   
    
    \item \textbf{Patent Generative Transformer} \cite{christofidellis2022pgt}: This model is a GPT-2 \cite{radford2019language} variant fine-tuned for multitasking with 11.6 million patents (1998-2020). The tasks included text infilling, text-to-text suggestions, and coherence checks. We used their HuggingFace model checkpoint \footnote{\url{https://huggingface.co/christofid/pgt}}, setting the maximum decoder length to 1024 and truncating input text to 256 words.

    \item \textbf{HUPD T5-Small} \cite{suzgun2022hupd}: Two separate T5-Small \cite{raffel2020exploring} models were fine-tuned on the HUPD dataset (2011-2016) for Description2Abstract and Claims2Abstract tasks. Claim-based summarization was observed to be more effective, which is also the task we tested in our work. We used the example codes from their GitHub \footnote{\url{https://github.com/suzgunmirac/hupd}}, maintaining default settings.

    \item \textbf{PatentGPT-J} \cite{ lee2023evaluating} pre-trained different sizes of GPT-J from scratch with 147B US patent data ranges from 1976-2020. Since the training data of claim generation is constructed within the schema of claim pairs as "claim n1<|dep|>claim n2" for explicating that claim n2 depends on claim n1. We kept this schema for each time the next claim generated depends on the previous claim. We tested the 1.6B model checkpoint from HuggingFace,\footnote{\url{https://huggingface.co/patent/PatentGPT-J-1.6B}} setting the maximum decoder length to 1024 and truncating the input claims to 512 words.
\end{itemize}

Additionally, we included potential LLMs such as Llama 2 \cite{touvron2023llama} \footnote{\url{https://huggingface.co/meta-llama/Llama-2-7b-hf}\url{https://huggingface.co/meta-llama/Llama-2-13b-hf}\url{https://huggingface.co/meta-llama/Llama-2-70b-hf}} and Falcon \cite{refinedweb}.\footnote{\url{https://huggingface.co/tiiuae/falcon-7b-instruct}\url{https://huggingface.co/tiiuae/falcon-40b-instruct}} For these models, we utilized text-generation-inference \footnote{\url{https://huggingface.co/docs/text-generation-inference/index}} for efficient inference, setting Falcon’s maximum length to 2048 and Llama2's to 4096. We also tested the GPT-3.5 \texttt{gpt-3.5-turbo-0613} version, setting the temperature to 0 to reduce randomness.

Uniform prompts were used for all three LLMs during inference to ensure a fair comparison:

\noindent
\textbf{Generate abstract given claims.} 
{\fontsize{8}{10}\selectfont 
\begin{lstlisting}[language=TeX, basicstyle=\ttfamily, breaklines=true, escapeinside={(*@}{@*)}]
Please draft a patent abstract from the provided claims. The abstract should concisely summarize the technical disclosure, enabling any reader to quickly understand the subject matter.
Claims: (*@\textcolor{blue}{\{claims\}}@*)
Abstract:
\end{lstlisting}}

\noindent
\textbf{Generate next dependent claim given previous claims.} 
{\fontsize{8}{10}\selectfont 
\begin{lstlisting}[language=TeX, basicstyle=\ttfamily, breaklines=true, escapeinside={(*@}{@*)}]
Please assist me in drafting the next DEPENDENT claim based on the provided patent claims below. This claim should be written in a dependent format, precisely specifying its dependency on one or more preceding claims. It should be legally sound, in line with patent claim drafting conventions, and use the existing claims as a basis for your draft. Ensure that the claim you draft is clearly and explicitly dependent on a previous claim.
Claims: (*@\textcolor{blue}{\{claims\}}@*)
\end{lstlisting}}

\noindent
\textbf{Generate next independent claim given previous claims.} 
{\fontsize{8}{10}\selectfont 
\begin{lstlisting}[language=TeX, basicstyle=\ttfamily, breaklines=true, escapeinside={(*@}{@*)}]
Please assist me in drafting the next INDEPENDENT claim in the series, directly following the provided patent claims below. This independent claim should be precise, legally sound, and in line with patent claim drafting conventions. Please continue the numbering scheme from the previous claims and ensure that this claim builds upon the previous claims logically.
Claims: (*@\textcolor{blue}{\{claims\}}@*)
\end{lstlisting}}

\section{Annotation Details}
\subsection{Data Statistics}

\begin{figure*}[ht]
    \centering
    \begin{subfigure}[t]{0.5\textwidth}
        \centering
        \includegraphics[height=2.5in]{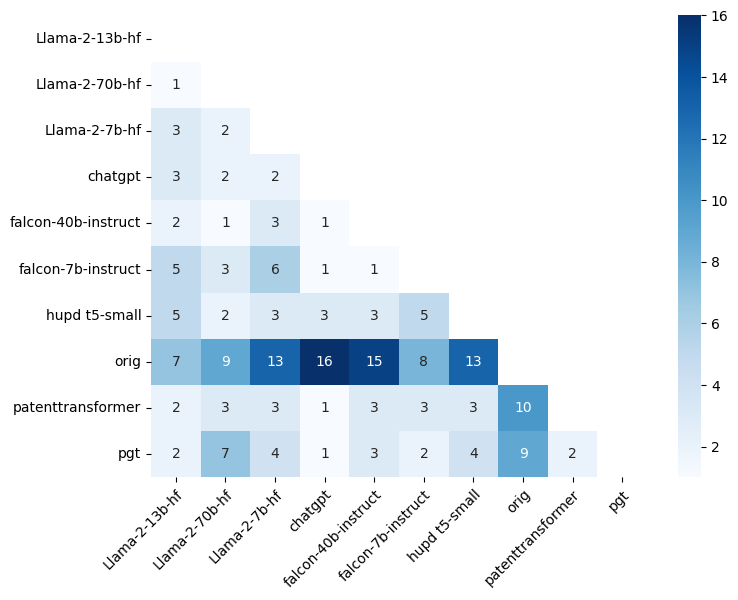}
        \caption{Selected model pairs for claims2abstract.}
    \end{subfigure}%
    ~ 
    \begin{subfigure}[t]{0.5\textwidth}
        \centering
        \includegraphics[height=2.5in]{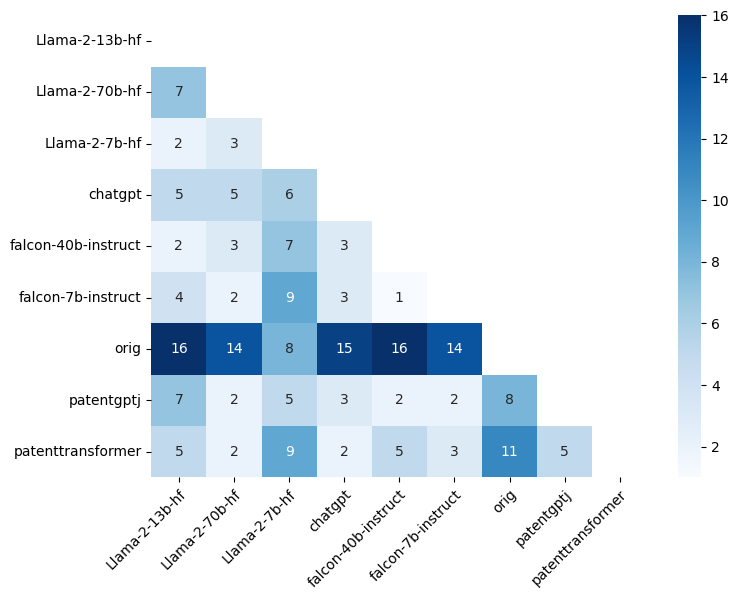}
        \caption{Selected model pairs for next claim generation.}
    \end{subfigure}
    \caption{Number of comparison model pairs selected for each task during human annotation.}
    \label{model1_model2}
\end{figure*}

Table~\ref{stat_data} presents statistical details of the 400 patents selected from the refined HUPD dataset, including the average number of claims, average word count in claims, and average word count in abstracts for each domain determined by the main IPC section of the respective patent data.

\begin{table}[ht!]
\small
\centering
\begin{tabular}{clrc}
\hline \textbf{domain} & \textbf{\# claims}& \textbf{\# words claims} & \textbf{\# words abstract}\\ \hline
 A & 15.2 & 952.86 & 101.72 \\
 B & 14.04 & 983.52 & 116.22 \\
 C & 17.36 & 1108.56 & 104.08 \\
 D & 14.86 & 740.22 & 106.28 \\
 E & 15.9 & 1059.1 & 123.26 \\
 F & 15.2 & 994.32 & 135.56 \\
 G & 14.7 & 1051.78 & 126.8 \\
 H & 15.36 & 1099.9 & 123.86 \\ \hline
\end{tabular}
\caption{\label{stat_data} Basic statistics of sampled patents. }
\end{table}

Additionally, Figure~\ref{model1_model2} illustrates the pairs of data samples chosen for our pairwise comparative analysis. Moving forward, our efforts will extend to annotating a broader range of examples produced by models across various domains. This expansion aims to deepen our understanding and provide a more comprehensive evaluation of model performance in diverse patent contexts.



\section{Metrics}
\label{Metrics}

Reflecting on our analyses, we explored different evaluation metrics for two patent generation tasks.

For abstract summarization, we employed two evaluation strategies. The first revolves around semantic similarity, assessing how closely the generated abstracts mirror the input claims in terms of meaning and context. The second strategy emphasizes the overlap of key technical features, focusing on the extent to which critical terms from the input claims are included in the output abstracts.

In the task of claim generation, we developed a metric to assess whether the generated claims adhere to some basic established guidelines for patent drafting, thus ensuring compliance. Additionally, we applied the same methodology used for evaluating claims-to-abstract semantic similarity. The details of these methods and their implementation are further discussed in the following subsections.

\subsection{Semantic Similarity Between Input Claims and Generated Abstracts}
\label{sec:semsim}

Inspired by the methodology used by Christofidellis et al. \cite{christofidellis2022pgt}, which evaluated model performance using semantic similarity via sentence transformers \cite{reimers2019sentence}. Our approach involved first fine-tuning BERT-for-patents \cite{SrebrovicYonamine2020}, a model specifically developed and trained for analyzing patent texts, on the patent IPC classification task. After fine-tuning, we used this specialized encoder to calculate the similarity between the input claims and the generated abstracts.  By adopting this strategy, we aim to provide a more specialized and relevant assessment tailored to patents. 

\subsubsection{Main IPC Classification}

The "Main IPC" of a patent corresponds to its primary IPC label, typically the first one assigned. For subclass-level main IPC classification (with over 600 labels in the label space), we fine-tuned a classifier using \texttt{bert-for-patents} \cite{SrebrovicYonamine2020}. The training data consisted of abstracts and claims of 1,338,054 patents filed in 2016 and 2017 from HUPD \cite{suzgun2022hupd}, and we tested the model on 63,862 patents filed in 2018. For the fine-tuning process, we utilized the PyTorch version checkpoint of \texttt{bert-for-patents} available on its HuggingFace page.\footnote{\url{https://huggingface.co/anferico/bert-for-patents}}

The \texttt{bert-for-patents} model was originally trained using meta-structures with special tokens like [abstract], [claims], [summary], [invention], etc., to indicate the corresponding section of the text. During the fine-tuning process, we also incorporated this information by adding the appropriate section token at the beginning of each input text. 
During the training phase, we configure the number of epochs to 3, set the learning rate to $1e-5$, and utilize a batch size of 64. To improve efficiency, we employ mixed precision training.

To evaluate the classifier's performance, we utilize data from HUPD filed in 2018. We test the fine-tuned model on three distinct test sets: one consisting solely of abstracts, another with only claims, and a final one with a mixture of both. We employ weighted precision, weighted recall, and weighted F1-score as metrics to measure the model's performance. Table~\ref{ipc_classification} displays the results for abstracts and claims, which exhibit similar performance levels compared to the combined dataset.

\begin{table}[ht!]
\small
\centering

\begin{tabular}{lccc}
\hline \textbf{Test set} & \textbf{Precision}&\textbf{Recall} & \textbf{F1-Score} \\ \hline
Abstract & 70.81 & 70.70 & 70.33 \\ \hline
Claims & 70.73 & 70.69 & 70.29 \\\hline
All & 70.76 & 70.70 & 70.32\\ \hline
\end{tabular}
\caption{\label{ipc_classification} Overall F1-score (weighted) of bert-for-patent for main IPC Classification Task. }
\end{table}



\subsubsection{Method}

To determine the relevance between the provided claims and the generated abstract, we utilize the encoder from the fine-tuned BERT-for-patent model and apply mean pooling to obtain vectors in the embedding space.


The relevance score is computed using the encoder model, denoted as $\Phi$, following this equation:
\begin{equation}
\text{Relevance Score}_i = \text{sim}(\Phi(x_i), \Phi(y_i))
\end{equation}
In this formula, $\text{sim}(\cdot)$ represents the cosine similarity function used to gauge the relevance between two embeddings. We employ cosine similarity as our metric of choice. $x_i$ and $y_i$ correspond to a pair of input claims and the generated abstract for a specific patent, labeled as patent $i$.

\subsection{Technical Features Coverage in Generated Abstracts}
\label{sec:term-coverage}
For a patent's abstract to be effective in search applications, it should encompass as many of the patent's technical features (terms) as possible. To aid in evaluating this aspect, we employed PyATE \cite{pyate}, specifically its \texttt{combo\_basic} algorithm,\footnote{\url{https://github.com/kevinlu1248/pyate}} for the task of term recognition. This algorithm was used to identify key terms within both the claims and the abstracts. We then measured term coverage using this formula:

\begin{equation} \label{eqn:term_coverage}
\text{Coverage}_i  = \frac{|\mathrm{U}(y_i)\cap\mathrm{U}(x_i)|}{|\mathrm{U}(x_i)|}
\end{equation}
Here, $x_i$ represents the input claims of the patent $i$, and $y_i$ denotes the generated abstract. And $\mathrm{U(x_i)}$ denotes the unique terms in the input claims $x_i$. This metric quantifies the extent to which the generated abstract captures the critical terms present in the original claims.




\subsection{N-grams Coverage in Generated Abstracts}

Similar to the method in \ref{sec:term-coverage}, the coverage rate of n-grams in generated abstracts was also analyzed. For this purpose, we considered n-grams ranging from n = 1 to 4 extracted by NLTK.

\subsection{Verification of Basic Rules for Generated Claims}

\begin{algorithm}[hbt!]
\small
\caption{Rule-based Evaluation Process for generated Claims}
\begin{algorithmic}[1]
    \State \textbf{Input:} input\_claims, generated\_claim, required\_dependency
    \State \textbf{Output:} score

    \State Extract numberings from input\_claims
    \State score $\gets$ 0
    \State total\_checks $\gets$ 4 \Comment{Total number of checks excluding distinctiveness}

    \If{generated\_claim is not distinctive}
        \State \Return 0 \Comment{Distinctiveness is a mandatory criterion}
    \EndIf

    \If{generated\_claim does not contain hallucinated repetitive content}
        \State score $\gets$ score + 1
    \EndIf

    \If{punctuation in generated\_claim is correctly placed}
        \State score $\gets$ score + 1
    \EndIf

    \If{numbering of generated\_claim follows consecutively after input\_claims}
        \State score $\gets$ score + 1
    \EndIf

    \If{dependency of generated\_claim is as required}
        \State score $\gets$ score + 1
    \EndIf

    \State score $\gets$ score / total\_checks \Comment{Normalize score}
    \State \Return score
\end{algorithmic}
\end{algorithm}

In the realm of patent claims, which are highly structured, rule-based evaluations can effectively identify errors in generated claims. We developed a set of rule-based checks to pinpoint various types of errors, such as grammatical inconsistencies, improper punctuation, sequential numbering errors, non-compliance with specified dependencies, lack of dependency clarity, and non-distinctive claim repetition.

Notably, non-distinctive claim repetition, where the content of a generated claim exactly mirrors an input claim, is treated as a critical error. If this error occurs, the evaluation process immediately returns a score of zero. For other errors, a point is added to the score for each rule the generated claim successfully adheres to. The final score is then normalized by dividing by the total number of checks, excluding the distinctiveness criterion.





\subsection{Semantic Similarity Between Input Claims and Generated Claim}
In assessing the semantic relevance between the given input claims and the generated subsequent claim, we employed the model trained as detailed in Section~\ref{sec:term-coverage}. The relevance score is calculated using the following formula:

\begin{equation}
\label{eqn:semsim2}
\begin{split}
&\text{Relevance Score}_i = \\
&\text{sim}(\Phi(c_{i1},...,c_{im}), \Phi(c_{i1},...,c_{im}, c'_{i(m+1)}))
\end{split}
\end{equation}

In this equation, $\Phi$ denotes the Bert-for-patents model (referenced in section~\ref{sec:semsim}). The sequence $c_{i1},...,c_{im}$ represents the input claims, while $c_{i1},...,c_{im}, c'_{i(m+1)}$ includes the input claims followed by the generated next claim. The function $\text{sim}(\cdot)$ computes the semantic similarity between the embedding of the concatenated input claims and the embedding of the input claims with the generated claim. 

We implemented this metric in two distinct manners. The first approach directly applies the calculation as defined in Equation~\ref{eqn:semsim2}. The second approach normalizes the similarity score by the score from the rule-based checker described in the previous subsection. This adjustment lowers the score for generated claims that fail to comply with patent drafting standards or the required dependency, offering a more comprehensive evaluation of the generated claim's quality.

\subsection{Other Studied Metrics}
\textbf{FactGraph} \cite{ribeiro2022factgraph} is specifically designed for evaluating factuality in document summarization tasks. This method utilizes advanced techniques for extracting Abstract Meaning Representation (AMR) graphs. The abstract and claims' graphs are firstly encoded using a graph encoder with structure-aware adapters. Additionally, text representations are generated using an adapter-based text encoder. These representations are then passed through a multilayer perceptron (MLP) to predict the factuality score. We obtained the implementation codes for the FactGraph method from the authors' GitHub repository.\footnote{FactGraph: \url{https://github.com/amazon-science/fact-graph}} 

\noindent
\textbf{QAFactEval} \cite{fabbri-etal-2022-qafacteval} is a QA-based metric for factual consistency evaluation in summarization, which is composed of four key components: 1) selection of answers for question generation, 2) question generation conditioned on these answers, 3) question answering based on the source document, and 4) evaluating the overlap between QA model output and selected answers. The codes we used were from the authors' GitHub.\footnote{QAFactEval: \url{https://github.com/salesforce/QAFactEval}}

\noindent
\textbf{EntityGrid} \cite{barzilay2008modeling} was grounded in the premise that the distribution of entities in locally coherent texts exhibits certain regularities, which can be formalized using entity-based theories of discourse. By leveraging these regularities, the proposed method can assess coherence as a machine-learning task. We modified the codes forked from coheoka.\footnote{EntityGrid: \url{https://github.com/ZoeYou/coheoka}}

\section{Examples of Typology of Errors}
\label{error_example}
In this section, we show examples of each type of error to articulate better the real scenario of mistakes made by models.

\noindent
\textbf{Claims to abstract generation errors.}

Given input claims as : 

\noindent
1. A computer system for selecting a version of a webpage to present to a user, the computer system comprising: one or more computer processors, one or more computer readable storage media, and program instructions stored on the one or more computer readable storage media for execution by at least one of the one or more processors, the program instructions comprising: program instructions to receive an indication of a user accessing the webpage, wherein the webpage includes a plurality of versions of the webpage, wherein the webpage is comprised of one or more modules, and wherein each version of the plurality of versions of the webpage comprises a different layout of the one or more modules; program instructions to retrieve a predefined goal associated with the webpage, wherein the predefined goal includes a higher number of sales, higher dollar amount per sale, length of time a user is on the webpage, usefulness of the webpage, and number of reviews written; program instructions to monitor usage information of the user accessing the plurality of versions of the webpage, based on the predefined goal associated with the webpage, wherein the usage information includes: cursor location, mouse clicks, idle time, HTML pages loaded, data accessed, widgets used, types of devices used to access the webpage, and presence of user interface artifacts; program instructions to generate a report that includes a collection of user characteristics, web analytics, and the monitored usage information; program instructions to store the monitored usage information of the plurality of versions of the webpage and the generated report; program instructions to receive a request to access the webpage from a device; program instructions to receive information about the device, wherein the information about the device includes: a device type, an Internet Protocol (IP) address, cookies, and a web browsing history; program instructions to access the monitored usage information of the plurality of versions of the webpage; program instructions to determine the version of the webpage of the plurality of versions of the webpage to present at the device, based on the information about the device, the predefined goal associated with the webpage, the monitored usage information, the generated report and user satisfaction information for the plurality of versions of the webpage, wherein the user satisfaction information comprises survey responses from the monitored accesses to the plurality of versions of the webpage; and program instructions to cause the determined versions of the webpage to be presented.

\onecolumn
\begin{center}
\small
\begin{longtable}{|p{0.17\textwidth}|p{0.83\textwidth}|}
\caption{Error typology of claims2abstract.} \label{error_type1} \\
\hline

\textbf{Error: }Grammatical Errors & \textbf{Description: }Occurrences of incorrect grammar, punctuation, or sentence structure, including repetitive or redundant sequences characteristic of language model outputs. \\ \hline
\textbf{Generated abstract} & In an approach for selecting a version of a webpage to present to \textcolor{red}{an} user, a processor receives a request to access a webpage from a device, wherein the webpage includes a plurality of versions of the webpage. A processor receives information about the device. A processor determines a version of the webpage to present, based on the information about the device \textcolor{red}{based on the information about the device based on the information about the device based on the information about the device} \\ \hline

\multicolumn{2}{|l|}{}  \\ \hline 

\textbf{Error: }Irrelevant Content & \textbf{Description: }Introducing content that deviates or digresses from the primary subject matter of the patent. \\ \hline
\textbf{Generated abstract} & In an advanced approach to enhancing user experience, the system not only selects a version of a webpage for presentation but also integrates \textcolor{red}{with a global weather forecasting service, providing real-time weather updates.} Upon receiving a request to access a webpage from a device, \textcolor{red}{where the webpage includes various versions, the processor unexpectedly engages in analyzing global culinary trends. It focuses on aggregating user preferences for different cuisines and correlates this with local restaurant recommendations, diverting significantly from the primary task of webpage version selection. The processor, while determining the webpage version based on the device information, simultaneously offers travel advice and booking options, extending far beyond the intended scope of webpage presentation.}\\ \hline

\multicolumn{2}{|l|}{}  \\ \hline 

\textbf{Error: }Incomplete Coverage & \textbf{Description: }Occurrences where the abstract omits essential components or concepts, failing to encapsulate all key points from the patent claims, especially the main (first independent) claim. \\ \hline
\textbf{Generated abstract} & In an approach for webpage presentation, a processor receives a request to access a webpage and information about the user's device. The processor then determines a version of the webpage to present based on the device information. This process simplifies the user's experience by selecting an appropriate webpage layout for their device.\\ \hline

\multicolumn{2}{|l|}{}  \\ \hline 

\textbf{Error: }Overly Wordy or Lengthy & \textbf{Description: }Abstracts falling into this error type are not succinct, containing unnecessary details. Jurisdictions often impose word limits on abstracts — for example, in many English-speaking countries, abstracts are typically restricted to 150 words. \\ \hline
\textbf{Generated abstract} & 
This invention involves a computer system comprising one or more computer processors, storage media, and a suite of complex program instructions. These instructions are meticulously designed to manage the intricate process of receiving an indication of a user accessing the webpage, which notably contains a variety of different versions. Each version uniquely features a distinct layout of several modules. The system is further programmed to retrieve a set of predefined goals associated with the webpage, encompassing diverse objectives like maximizing sales, increasing the average dollar amount per sale, enhancing the duration a user spends on the webpage, improving the overall usefulness of the webpage, and encouraging the generation of a larger number of reviews.

Furthermore, the computer system is engineered to monitor an array of user-specific usage information while they navigate through the different versions of the webpage. This monitoring is intricately aligned with the predefined goals and includes tracking various elements such as cursor location, mouse clicks, periods of idle time, the types of HTML pages loaded, data accessed, widgets utilized, types of devices used for accessing the webpage, and the presence of any user interface artifacts. In addition to these functionalities, the system is adept at generating comprehensive reports that include a collection of user characteristics, detailed web analytics, and the monitored usage information.

The system's capabilities extend to storing the monitored usage information of the various versions of the webpage alongside the generated report. When a request to access the webpage is received from any device, the system promptly acquires specific information about the device, including but not limited to the device type, the Internet Protocol (IP) address, cookies, and a detailed web browsing history. Post this, it accesses the stored monitored usage information of the webpage's multiple versions.

Based on a thorough analysis of the device information, the predefined goals associated with the webpage, the meticulously monitored usage information, the detailed generated reports, and the user satisfaction information—which encompasses survey responses from users who have previously accessed the different versions of the webpage—the system intelligently determines the most suitable version of the webpage to be presented on the requesting device. Finally, it takes the necessary steps to ensure that the determined version of the webpage is accurately and efficiently presented to the user.\\ \hline

\multicolumn{2}{|l|}{}  \\ \hline 

\textbf{Error: }Contradictory Information & \textbf{Description: }Instances when the abstract introduces factual details that contradict the content found in the original claims. \\ \hline
\textbf{Generated abstract} & In a novel system for webpage version selection, \textcolor{red}{a single, static version of a webpage is presented to all users}, streamlining user experience. Upon receiving a device's access request, the system evaluates the \textcolor{red}{user's geographic location and the time} of access to determine webpage presentation. \\ \hline

\multicolumn{2}{|l|}{}  \\ \hline 

\textbf{Error: }Unclarity & \textbf{Description: }The abstract contains vague or ambiguous descriptions, making it difficult to grasp the intended message or details. \\ \hline
\textbf{Generated abstract} &  In a system for choosing a webpage version, a processor processes a webpage access request, where the webpage includes multiple versions. The processor analyzes \textcolor{red}{some device information}. It then selects a webpage version to display, based on the device information and a \textcolor{red}{general goal} associated with the webpage. Subsequently, the processor facilitates the presentation of a version of the webpage.\\ \hline

\multicolumn{2}{|l|}{}  \\ \hline 

\textbf{Error: }Ineffective Summarization & \textbf{Description: }Relates to abstracts that inadequately summarize the invention, often replicating one of the claims verbatim instead of providing a concise and comprehensive overview of the entire patent. \\ \hline
\textbf{Generated abstract} & (A direct copy of first claim or several claims.)\\ \hline
\end{longtable}
\end{center}

\noindent
\textbf{Next claim generation errors.}

\begin{center}
\small
\begin{longtable}{|p{0.2\textwidth}|p{0.8\textwidth}|}
\caption{Error typology of next claim generation.} \label{error_type2} \\
\hline

\multicolumn{2}{|l|}{\textbf{Grammatical Errors}} \\ \hline
\textbf{Error: } Grammatical Inaccuracy & \textbf{Description: } Misuse of grammar and hallucinated repetitive sequences produced by language models. \\ \hline
\textbf{Input claims} & 1. A vaccine composition comprising an antigen or antigen composition, and a compound of formula (I) or a salt thereof: wherein: R1 is n-C3-6alkyl or C1-2alkoxyC1-2alkyl-; each R2 independently represents halo, OH or C1-3alkyl; m is an integer having a value of 4, 5, 6 or 7; n is an integer having a value of 0, 1, 2 or 3; p is an integer having a value of 0, 1 or 2. \\ \hline
\textbf{Generated claim} & 2. The vaccine composition of claim 1, wherein the compound of formula (I) is 4-\textcolor{red}{((4-((4-((4-((4-((4-((4-((4-((4-((4-((4-((4-((4-((4-((4-((4-((4-((4-((4-((4-((4-((4-((4-((4-((4-((4-((4-((4-((4-((4-((4-((4-((4-((4-((4-((4-((4-((4-((4-((4-((4-((4-((4-((4-((4-((4-((4-((4-((4-((4-((4-((4-((4-((4-((4-((4-((4-((4-((4-((4-((4-((4-((4-((4-((4-((4-((4-((4-((4-((4-((4-((4-((4-((4-((4-((4-((4-((4-((4-((4-((4-((4-((4-((4-((4-((4-((4-((4-((4-((4-((4-((4-((4-((4-((4-((4-((4-((4-((4-((4-((4-((4-((4-((4-((4-((4-((4-((4-((4-((4-((4-((4-((4-((4-((4-((4-((4-((4-((4-((4-((4-((4-((4-((4-((4-((4-((4-((4-((4-((4-((4-((4-((4-((4-((4-((4-((4-((4-((4-((4-((4-((4-((4-((4-((4-((4-((4-((4-((4-((4-((4-((4-((4-((4-((4-((4-((4-((4-((4-((4-((4-((4-((4-((4-} \\ \hline \hline

\textbf{Error: } Punctuation Discrepancy & \textbf{Description: }Incorrect or inconsistent use of punctuation marks, deviating from standard patent drafting conventions. \\ \hline
\textbf{Input claims} & 1. A multiphasic pulse generator, comprising: a subsystem that generates a first phase of a pulse, the subsystem having a power source and an energy reservoir, wherein the first phase has a shape and is one of a positive phase of the pulse and a negative phase of the pulse; a separate subsystem that generates a second phase of the pulse, the separate subsystem having a second power source and a second energy reservoir, wherein the second phase is shaped independently from the shape of the first phase and is an opposite polarity phase to the first phase; and a switch element that switches between the subsystem and the separate subsystem to generate a therapeutic pulse having at least one positive phase and at least one negative phase. \\ \hline
\textbf{Generated claim} & 2. The device as claimed in claim 1\textcolor{red}{\st{,}} further including means for controlling said switching by varying voltage applied across said switch element. \\ \hline
\multicolumn{2}{|l|}{}  \\ \hline 
\multicolumn{2}{|l|}{\textbf{Formatting Errors}} \\ \hline
\textbf{Error: } Claim Numbering Error & \textbf{Description: }Incorrect or inconsistent numbering of claims.\\ \hline
\textbf{Input claims} & 1. A coin processing device comprising: a coin passage through which an inserted coin passes; a material detection sensor which includes a first coil and a second coil facing each other with the coin passage interposed therebetween; an outer diameter detection sensor which includes a ring-shaped third coil that surrounds the first coil and a ring-shaped fourth coil that surrounds the second coil, the third coil and the fourth coil facing each other with the coin passage interposed therebetween; a first oscillation circuit which is connected to the material detection sensor and oscillates a first oscillation signal in an individual connection state, and is connected to the material detection sensor and the outer diameter detection sensor that are connected in series and oscillates the first oscillation signal in a series connection state; a second oscillation circuit which is connected to the outer diameter detection sensor and oscillates a second oscillation signal in the individual connection state; a switching unit which switches the individual connection state and the series connection state; and a coin identification unit which detects an outer diameter of the coin using the second oscillation signal in the individual connection state or the first oscillation signal in the series connection state and identifies the coin based on the outer diameter. \\ \hline
\textbf{Generated claim} & \textcolor{red}{1.} A method for processing a coin, comprising: passing a coin through a coin passage of a coin processing device; detecting the coin's outer diameter using a first oscillation signal in an individual connection state; and processing the coin based on the detected outer diameter. \\ \hline \hline 

\textbf{Error: } Preamble Inconsistency Error & \textbf{Description: }Inaccurate reflection of subject matter in the preamble, disrupting the conceptual flow between independent and dependent claims.\\ \hline
\textbf{Input claims} &  1. \textcolor{blue}{An unmanned aerial vehicle (UAV)} configured to guide a target, said UAV comprising: one or more vision sensors configured to capture at least one image of the target; one or more processors configured to, individually or collectively, recognize the target from the image of the target; an attractor display mechanism configured to display, without human aid or intervention, an attractor to the target when the target is recognized; and one or more propulsion units configured to permit flight of the UAV while the attractor is displayed to the target. \\ \hline
\textbf{Generated claim} & 2. \textcolor{red}{The UAVs} as recited in claim 1 wherein each sensor comprises a camera that captures images using infrared light. \\ \hline \hline 

\textbf{Error: } Transitional Phrase Error & \textbf{Description: } Improper use of transitional phrases, impacting the scope of the claim. \\ \hline
\textbf{Input claims} & 1. A portable, personal storage and carrying case for an e-liquid e-cigarette PV in which the case includes: (a) an electrical power source for re-charging a rechargeable battery in the PV; (b) a user-replaceable reservoir for holding e-liquid; and (c) an electrical or electronic pump adapted to transfer e-liquid from the reservoir to a chamber in the PV, the pump delivering a pre-defined or variable quantity of e-liquid from the reservoir; and in which the case is configured to implement a nicotine reduction or cessation program. \\ \hline
\textbf{Generated claim} & 2. The case of claim 1, \textcolor{red}{not only including} an electrical power source for re-charging a rechargeable battery in the PV, \textcolor{red}{but also} the case is peculiarly configured to automatically fully re-fill the PV with e-liquid and fully charge the PV so that the PV is in a fully re-filled and re-charged state when it is removed from the case. \\ \hline \hline 

\textbf{Error: } Claim Body Disconnection & \textbf{Description: } Presence of fewer than two elements or a lack of a coherent, logical connection between listed elements in the claim body. \\ \hline
\textbf{Input claims} & 1. A portable, personal storage and carrying case for an e-liquid e-cigarette PV in which the case includes: (a) an electrical power source for re-charging a rechargeable battery in the PV; (b) a user-replaceable reservoir for holding e-liquid; and (c) an electrical or electronic pump adapted to transfer e-liquid from the reservoir to a chamber in the PV, the pump delivering a pre-defined or variable quantity of e-liquid from the reservoir; and in which the case is configured to implement a nicotine reduction or cessation program. \\ \hline
\textbf{Generated claim} & 2. The case of claim 1, wherein the case is configured to automatically fully re-fill the PV with e-liquid, and further comprising \textcolor{red}{a secondary storage compartment}. \\ \hline
\multicolumn{2}{|l|}{}  \\ \hline 

\multicolumn{2}{|l|}{\textbf{Dependency Errors}} \\ \hline
\textbf{Error: } Non-compliant Dependency with instruction & \textbf{Description: } Dependency of the claims not matching the required dependency as instructed. \\ \hline
\textbf{Input claims} & 1. A coin processing device comprising: a coin passage through which an inserted coin passes; a material detection sensor which includes a first coil and a second coil facing each other with the coin passage interposed therebetween; an outer diameter detection sensor which includes a ring-shaped third coil that surrounds the first coil and a ring-shaped fourth coil that surrounds the second coil, the third coil and the fourth coil facing each other with the coin passage interposed therebetween; a first oscillation circuit which is connected to the material detection sensor and oscillates a first oscillation signal in an individual connection state, and is connected to the material detection sensor and the outer diameter detection sensor that are connected in series and oscillates the first oscillation signal in a series connection state; a second oscillation circuit which is connected to the outer diameter detection sensor and oscillates a second oscillation signal in the individual connection state; a switching unit which switches the individual connection state and the series connection state; and a coin identification unit which detects an outer diameter of the coin using the second oscillation signal in the individual connection state or the first oscillation signal in the series connection state and identifies the coin based on the outer diameter. \\ \hline
\textbf{Generated claim} & 1. A method for processing a coin, comprising: passing a coin through a coin passage of a coin processing device; detecting the coin's outer diameter using a first oscillation signal in an individual connection state; and processing the coin based on the detected outer diameter. (\textcolor{red}{For this specific input claim, we asked for generating a next dependent claim.}) \\ \hline \hline 

\textbf{Error: } Dependency Clarity Error & \textbf{Description: } Utilization of unclear multiple dependencies or an incorrect singular dependency. \\ \hline
\textbf{Input claims} & 1. A method comprising: identifying a set of uncompressed pages in a primary memory storage; assigning to each uncompressed page in the set of uncompressed pages one or more corresponding classes from a pre-determined set of classes; compressing a plurality of pages in the set of uncompressed pages, a first set of pages of the plurality of pages having been assigned to a first class are compressed into a single logical page, a second set of pages of the plurality of pages being assigned to a corresponding class different than the first class are compressed into a separate logical page different than the single logical page; determining a page weight factor for the single logical page; paging in the single logical page into the primary memory storage; prompting a hypervisor to steal the single logical page, based, at least in part, on the page weight factor; wherein: the page weight factor is made up of component weight factors assigned individually to each page within the single logical page; and paging in the single logical page causes only a set of compressed pages assigned to the same class to be paged into the primary memory storage. \\ \hline
\textbf{Generated claim} & The method of claim \textcolor{red}{22}, wherein, in response to paging the single logical page into primary memory, determining a page weight factor includes: determining an average page size of the set of uncompressed pages as a result of paging using the page weight factor. \\ \hline \hline 

\textbf{Error: } Broad Scope Dependent Claims & \textbf{Description: } Dependent claims that insufficiently narrow the scope of the independent claim they depend on. \\ \hline
\textbf{Input claims} & 1. A camera mounting assembly for adjustably mounting a camera on a vertical support comprising: (a) a base member having an upper portion configured to support a lower surface of the camera and a lower portion forming an opposite pair of lower inclined edges; (b) a side member forming an opposite pair of side inclined edges; and (c) a fastening element configured to detachably fasten together said side member and said base member such that when the lower surface of the camera is supported by the base member, at least a portion of the side member is aligned with a side surface of the camera and configured such that a pair of moveable jaws of a quick-release mechanism is selectively engageable with the lower inclined edges and side inclined edges to adjustably mount the camera in a landscape orientation and a portrait orientation, respectively, while maintaining the camera in a generally centered position over the vertical support, said fastening element comprising a rotatable member accessible from a bottom surface of said base member when said lower surface of said camera is supported by said base member where rotation of said rotatable member detachably interconnects said side member to said base member when said camera is supported by said base member, said rotatable member movable in a first manner that permits said side member to slide with respect to said base member in such a manner to modify the spacing between said side member and said side surface of said camera when said lower surface of said camera is said supported by said base member, said rotatable member movable in a second manner that prevents said side member to slide with respect to said base member in such a manner to maintain the spacing between said side member and said side surface of said camera when said lower surface of said camera is said supported by said base member, said base member being free from extending along a substantial portion of said side surface of said camera when said camera is supported by said base member and said side member is detached from said base member; (d) wherein said side member includes a leg portion that engages with said base member and said fastening element detachably fastens together said leg portion of said side member and said base member.\newline 2. The camera mounting assembly of claim 1 wherein the side member includes a side arm and a lower arm in a generally L-shaped arrangement, the side arm being aligned with the side surface of the camera when the lower surface of the camera is supported by the base member.\newline 3. The camera mounting assembly of claim 2 wherein the lower arm has a length \textcolor{blue}{shorter than the length of both the base member and the side arm.} \\ \hline
\textbf{Generated claim} & 4. The camera mounting assembly of claim 3 wherein the lower arm has a length \textcolor{red}{shorter than the length of the base member.} \\ \hline \hline 

\textbf{Error: } Insufficient Differentiation of Independent Claims & \textbf{Description: } Independent claims that cover the same or similar scope as previous claims. \\ \hline
\textbf{Input claims} & 1. A drill bit, comprising: a bit body having an axis defining an axial center of the bit body, at least one spindle, and at least one fixed blade extending in an axial direction downwardly from the bit body; at least one roller cone mounted on the at least one spindle of the bit body; at least one rolling-cutter cutting element arranged on the roller cone and radially spaced apart from the axial center; a plurality of fixed cutting elements arranged on the at least one fixed blade, at least one of the fixed cutting elements of the plurality of fixed cutting elements being located near the axial center of the bit body and adapted to cut formation at the axial center; and a bearing assembly between the at least one spindle and the at least one roller cone, the bearing assembly comprising a plurality of polycrystalline diamond compact bearing elements.\newline\textcolor{blue}{2. A hybrid drill bit for use in drilling through subterranean formations, the hybrid drill bit comprising: a shank having a longitudinal centerline and adapted to be coupled to a drilling string; at least one fixed blade extending from the shank, the at least one fixed blade comprising at least one cutting element extending from a surface of the at least one fixed blade; and at least two rolling cutter legs extending downwardly from the shank, each of the legs of the at least two rolling cutter legs comprising: a cantilevered bearing shaft extending inwardly toward the longitudinal centerline and downwardly; a roller cone mounted for rotation on the cantilevered bearing shaft, the roller cone adapted to rotate about the cantilevered bearing shaft, the roller cone comprising a plurality of cutting elements extending from an external surface of the roller cone; and a bearing assembly between the cantilevered bearing shaft and the roller cone, the bearing assembly comprising a plurality of polycrystalline diamond compact bearing elements coupled to one or both of the cantilevered bearing shaft and the roller cone.\newline 3. The hybrid drill bit of claim 2, wherein the bearing assembly of at least one of the at least two rolling cutter legs comprises a pin extending from the respective rolling cutter leg and an external bearing assembly mounted on the pin.} \\ \hline
\textbf{Generated claim} & 4. \textcolor{red}{A hybrid drill bit for use in drilling through subterranean formations, the hybrid drill bit comprising: a shank having a longitudinal centerline and adapted to be coupled to a drilling string; at least one fixed blade extending from the shank, the at least one fixed blade comprising at least one cutting element extending from a surface of the at least one fixed blade; and at least two rolling cutter legs extending downwardly from the shank, each of the legs of the at least two rolling cutter legs comprising: a cantilevered bearing shaft extending inwardly toward the longitudinal centerline and downwardly; a roller cone mounted for rotation on the cantilevered bearing shaft, the roller cone adapted to rotate about the cantilevered bearing shaft, the roller cone comprising a plurality of cutting elements extending from an external surface of the roller cone; and a bearing assembly between the cantilevered bearing shaft and the roller cone, the bearing assembly comprising a plurality of polycrystalline diamond compact bearing elements coupled to one or both of the cantilevered bearing shaft and the roller cone; wherein the bearing assembly of at least one of the at least two rolling cutter legs comprises a pin extending from the respective rolling cutter leg and an external bearing assembly mounted on the pin.} \\ \hline
\multicolumn{2}{|l|}{}  \\ \hline

\multicolumn{2}{|l|}{\textbf{Clarity Errors}} \\ \hline
\textbf{Error: } Vagueness & \textbf{Description: } Usage of ambiguous, vague, or relative terms or expressions that render the claim's scope indefinite. \\ \hline
\textbf{Input claims} & 1. A method of treating hyperalgesia associated with complex regional pain syndrome, comprising parenterally administering neridronic acid in a salt or an acid form to a human being suffering from hyperalgesia associated with complex regional pain syndrome.\newline2. The method of claim 1, wherein a total of about 200 mg to about 500 mg of the neridronic acid is administered parenterally to the human being. \\ \hline
\textbf{Generated claim} & 3. The method of claim 1, wherein the neridronic acid is administered in a \textcolor{red}{salt form}. \\ \hline \hline 

\textbf{Error: } Antecedent Reference Errors & \textbf{Description: }Failure to provide a clear antecedent basis for each term. \\ \hline
\textbf{Input claims} & 1. A hip protector system comprising: a pouch configured to be worn around a waist of a user, wherein the pouch includes an inflatable airbag; a proximity sensor configured to measure a distance from the user's pelvis to a point on the ground surface; a gyro configured to provide spatial orientation of the waist plane of the user; an accelerometer configured to determine the vertical acceleration of the waist plane of the user; a micro controller unit operatively connected to said proximity sensor, said gyro, and said accelerometer; wherein the micro controller unit is configured to: receive a vertical acceleration direction from the accelerometer; receive a proximity measurement distance signal, from the proximity sensor; receive or compute spatial orientation relative to the vertical direction based on signals provided by the gyro; calculate a height based on said proximity measurement distance signal and said spatial orientation; compute a first value of downward velocity based on a change of the height in time; compute a second value of downward velocity based on the vertical acceleration integrated over time; and correlate the first value of downward velocity and the second value of downward velocity to validate a true height in time from the waist plane of the user to the ground surface, wherein the airbag is inflated responsive to the true height from the waist plane of the user to the ground surface to protect the user.\newline 2. \textcolor{blue}{The hip protector system} of claim 1, wherein the inflatable airbag includes a first airbag and a second airbag, wherein the first airbag is configured to be wrapped against the right part of the waist of the user, and the second airbag is configured to be wrapped against the left part of the waist of the user. \\ \hline
\textbf{Generated claim} & 3. \textcolor{red}{A hip protector system} according to claim 2, wherein the pouch further includes at least one inflatable band configured to secure the lower portion of the body of the user to the pouch. \\ \hline \hline 

\textbf{Error: } Terminological Inconsistency & \textbf{Description: } Use of multiple terms or different reference numerals for the same element. \\ \hline
\textbf{Input claims} & 1. A board game system comprising: a plurality of \textcolor{blue}{blocks} configured to form a changing passageway based in part in the orientation of the plurality of blocks with respect to each other; the plurality of blocks including; a base with planar sidewalls and a level bottom surface, projections extending away from an upper surface of the base; a channel positioned between the projections, the channel extending from a first corner of the block to a second corner of the block, wherein sidewalls of the projections define the channel; grooves positioned at a third corner of the block and a fourth corner of the block, wherein upper surface grooves are positioned at the same vertical offset of an upper surface of the channel.\newline2. The system of claim 1, wherein a distance from the first corner of the block to a first end of a first sidewall of a first projection is the same distance from the third corner of the block to a second sidewall of the first projection.\newline3. The board game system of claim 1, wherein the plurality of blocks includes nine blocks with nine independent channels, wherein each of the nine independent channels is part of the passageway. \\ \hline
\textbf{Generated claim} & 4. The board game system of claims 1 or 2 further comprises means for aligning two adjacent ones of \textcolor{red}{said boards} when they are placed side by side such that one of them can be moved into position while another has its opposite edge aligned so as not interfere with its movement.\\ \hline \hline 

\textbf{Error: } Wishful Claiming & \textbf{Description: } Claims that express objectives without concrete methods, leading to speculative or abstract language. \\ \hline
\textbf{Input claims} & 1. An image forming apparatus for forming images on media through an image forming operation, comprising a carrying part that carries the media along a carrying path for the image forming operation, each of the media has a leading edge at a downstream and a trailing edge at an upstream in the carrying path, a supply part that is located at an upstream in the carrying part and forwards the media to the carrying part piece by piece, wherein when a preceding medium, which is one of the media, is detected to have been fed to the carrying part, a succeeding medium, which is another one of the media, is next fed to the carrying part, and a carrying controller that stops the preceding medium at a stop position of the media before the preceding medium is carried to the image forming process, the stop position being defined as a position where the leading edge of the media is stopped, wherein the carrying controller obtains a medium length of the preceding medium in the carrying path, and changes the stop position of the preceding medium based on the medium length such that the trailing edge of the preceding medium is maintained distant from the leading edge of the succeeding medium while the preceding medium stops at the stop position. \\ \hline
\textbf{Generated claim} & 2. The image forming apparatus according to claim 1, wherein the apparatus \textcolor{red}{enhances the overall efficiency and quality of image formation on the media, ensuring faster processing times and improved clarity of the images formed, while also reducing the energy consumption and operational costs.} \\ \hline
\multicolumn{2}{|l|}{}  \\ \hline

\multicolumn{2}{|l|}{\textbf{Brevity Errors}} \\ \hline
\textbf{Error: } Verbose Redundancy & \textbf{Description: } Excessive wordiness without adding substantive content. \\ \hline
\textbf{Input claims} & 1. A system for providing health-related information about a person comprising: a two-dimensional barcode linked to the health-related information, \textcolor{blue}{wherein the two-dimensional barcode is configured to be scanned}; and an algorithm configured to program a microprocessor to process the two-dimensional barcode upon receiving the two-dimensional barcode obtained by a scanner.\newline2. The system of claim 1, wherein the scanner comprises the microprocessor.\newline3. The system of claim 1, wherein the scanner comprises a wireless device configured for optical detection. \\ \hline
\textbf{Generated claim} & 4. The system of claim 1, \textcolor{red}{wherein the two-dimensional barcode is configured to be scanned} by a scanner. \\ \hline \hline 

\textbf{Error: } Sub-Optimal Claim Structure & \textbf{Description: } Claims with complex language that could be more clearly expressed as multiple, simpler claims. \\ \hline
\textbf{Input claims} & 1. A biological fluid sample analysis cartridge, comprising: a housing; a fluid module having a sample acquisition port and an initial channel, which fluid module is connected to the housing, and which initial channel is in fluid communication with the acquisition port; and an imaging tray having an analysis chamber, which tray is selectively positionable relative to the housing in an open position and a closed position, and in the closed position, the analysis chamber is in fluid communication with the initial channel.\newline2. The cartridge of claim 1, wherein the imaging tray is selectively lockable in the closed position, in which position it is disposed within the housing.\newline3. The cartridge of claim 1, further comprising a magnetically actuable latch selectively operable to lock or unlock the imaging tray in the closed position. \\ \hline
\textbf{Generated claim} & 4. The cartridge of claim 1, wherein the imaging tray is selectively lockable in the closed position, in which position it is disposed within the housing, \textcolor{red}{and wherein the imaging tray is further operable to be selectively positioned relative to the housing in an open position and a closed position, in the closed position, the imaging tray is in fluid communication with the initial channel.} \\ \hline
\multicolumn{2}{|l|}{}  \\ \hline 

\multicolumn{2}{|l|}{\textbf{Content Relevance Errors}} \\ \hline
\textbf{Error: } Irrelevant Matter Introduction & \textbf{Description: } Introduction of matter unrelated to the disclosed embodiments, potentially broadening the claim beyond the invention's scope. \\ \hline
\textbf{Input claims} & 1. A method for using a resource by one or more applications, the resource comprising multiple resource components that are individually accessed and controlled by an operating system for being used by the one or more applications, each of the resource components is tagged using a first tag, a second tag, or a third tag, and each of the resource components is capable of being used by the one or more applications for a first purpose and a second purpose, for use with a request from an application by an operating system to use two resource components respectively for the first and second purposes, the method comprising the steps of: determining if a resource component associated with the first tag or with the second tag is available for use; responsive to the determining, notifying the application if no resource component in the resource is associated with the first tag or with the second tag; determining, by the operating system, if a first resource component associated with the first tag is available in the resource; if a first resource component associated with the first tag is available, then: selecting the first resource component associated with the first tag; using the selected first resource component by the application for the first purpose; and tagging the first resource component with the third tag; determining, by the operating system, if a third resource component associated with the first tag is available in the resource; if a third resource component associated with the first tag is available, then: selecting the third resource component associated with the first tag; using the third resource component by the application for the second purpose; and tagging the third resource component with the second tag. \\ \hline
\textbf{Generated claim} & 2. \textcolor{red}{The computer-readable medium having stored thereon instructions which when executed perform the method according to claim 1 further including storing information about resources currently assigned as either private or public depending on whether they were previously allocated to other users.} \\ \hline
\multicolumn{2}{|l|}{}  \\ \hline

\multicolumn{2}{|l|}{\textbf{Effectiveness Error}} \\ \hline
\textbf{Error: } Contradictory Claims & \textbf{Description: } Claims that conflict with previous claims or do not follow a logical flow themselves. \\ \hline
\textbf{Input claims} & 1. A fire monitoring system, comprising: a smoke detector including a light emitting element and a light receiving element provided in a smoke detection chamber, the smoke detector being configured to output a detection value of the light receiving element corresponding to a smoke density in the smoke detection chamber; a fire alarm control unit configured to receive output from the smoke detector; a reference value storage unit configured to store a reference value, the reference value being the detection value of the light receiving element when the smoke density is zero; a first correction unit configured to obtain a first corrected value by multiplying a difference value between the reference value and the detection value of the light receiving element by a first correction coefficient; a first conversion unit configured to convert the first corrected value into a first smoke density; and a fire determination unit configured to determine occurrence of a fire event based on a result of comparison between the first smoke density and a fire threshold value, wherein the first correction coefficient is set on an \textcolor{blue}{increase} side corresponding to an increase in a rate of change of the reference value with respect to an initial reference value, the initial reference value being an initial value of the reference value, and wherein an upper limit value is set for the first correction coefficient. \\ \hline
\textbf{Generated claim} & 2. The fire monitoring system according to claim 1, wherein the first correction coefficient is set on a \textcolor{red}{decrease} side corresponding to a decrease in the rate of change of the reference value with respect to the initial reference value. \\ \hline \hline
\textbf{Error: } Non-Distinctive Claim Repetition & \textbf{Description: } Claims that lack effectiveness, primarily repeating content from earlier claims without adding new scope or detail. \\ \hline
\textbf{Input claims} & 1. A biological fluid sample analysis cartridge, comprising: a housing; a fluid module having a sample acquisition port and an initial channel, which fluid module is connected to the housing, and which initial channel is in fluid communication with the acquisition port; and an imaging tray having an analysis chamber, which tray is selectively positionable relative to the housing in an open position and a closed position, and in the closed position, the analysis chamber is in fluid communication with the initial channel.\newline\textcolor{blue}{2. The cartridge of claim 1, wherein the imaging tray is selectively lockable in the closed position, in which position it is disposed within the housing.}\newline3. The cartridge of claim 1, further comprising a magnetically actuable latch selectively operable to lock or unlock the imaging tray in the closed position. \\ \hline
\textbf{Generated claim} & \textcolor{red}{4. The cartridge of claim 1, wherein the imaging tray is selectively lockable in the closed position, in which position it is disposed within the housing.} \\ \hline

\hline
\end{longtable}
\end{center}

\section{Annotation Interface}
\begin{figure*}[b!]
    \centering
    \includegraphics[width=1\textwidth]{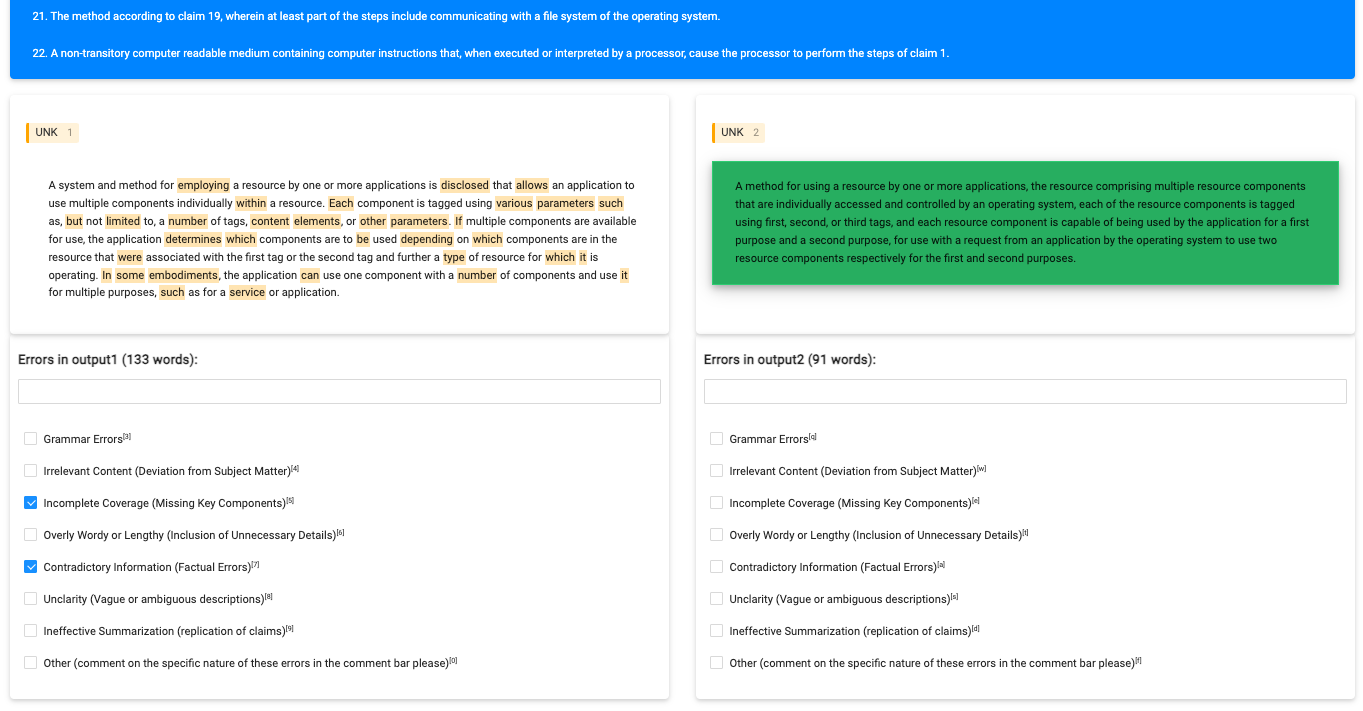}
    \caption{Annotation interface for claims-to-abstract task pairwise comparisons and error annotations.}
    \label{fig:interface1}
\end{figure*}

\begin{figure*}[b!]
    \centering
    \includegraphics[width=1\textwidth]{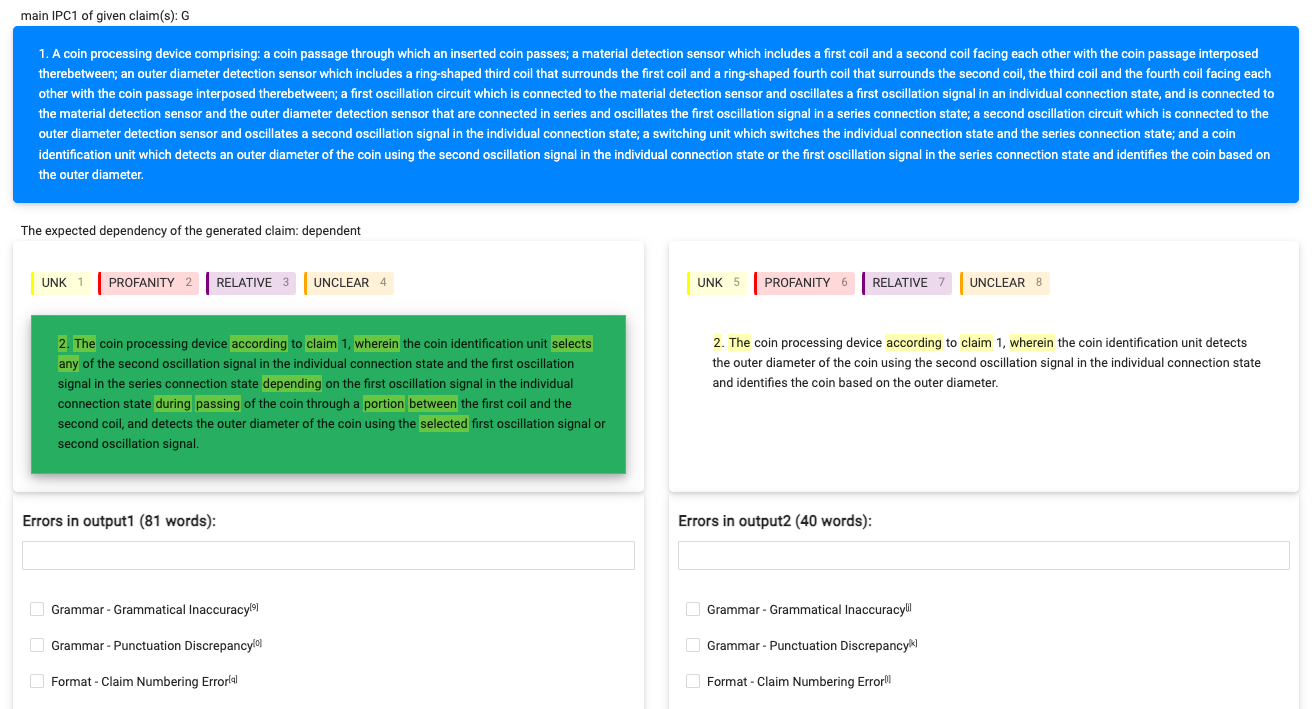}
    \caption{Annotation interface for next-claim-generation task pairwise comparisons and error annotations.}
    \label{fig:interface2}
\end{figure*}

We utilized Label Studio\footnote{https://labelstud.io/} to create the annotation interface for our study. Screenshots of this interface for the studied task are shown in Figure~\ref{fig:interface1} and~\ref{fig:interface2}. The preprocessed annotation data files and platform settings will be available.

In the annotation process, annotators were presented with the two outputs (output1 and output2) for each set of input claims. They were tasked with selecting the preferred output. The patent domain was indicated for context at the top, and words in the outputs not present in the input texts were highlighted to assist the annotators in their evaluation. Additionally, a comment section was provided below each output, allowing annotators to note any observations or justifications for their choices.

Beyond pairwise comparisons, annotators were also instructed to identify any errors present in the outputs. They were furnished with detailed descriptions and examples of possible error types, similar to those listed in Table~\ref{error_type1} and Table~\ref{error_type2}. This comprehensive approach was designed to ensure a thorough and nuanced assessment of each model's performance in generating patent texts.

\newpage

\end{document}